\documentclass[preprint]{elsarticle}

\usepackage{lineno,hyperref}
\usepackage{graphicx}
\usepackage{hyperref}
\usepackage{soul}
\usepackage{xcolor}
\usepackage{xargs}
\usepackage{rotating}
\usepackage{amsmath}
\usepackage{subfig}
\usepackage{natbib}
\usepackage{setspace}
\usepackage{etoolbox}
\modulolinenumbers[5]

\newcommand{\rpm}{\raisebox{.2ex}{$\scriptstyle\pm$}}

\newcommandx{\hlc}[3][1=zzz, 2=yellow]{{%
		\colorlet{foo}{#2}%
		\sethlcolor{foo}\hl{#3}}%
}

\DeclareMathOperator*{\argmax}{arg\,max}

\journal{Journal of \LaTeX\ Templates}

\AtBeginEnvironment{tabular}{\singlespacing}% Single spacing in tabular environment
\begin{document}

\begin{frontmatter}

\title{Pitfalls of Assessing Extracted Hierarchies for Multi-Class Classification}


\author[mymainaddress]{Pablo del Moral}\corref{mycorrespondingauthor}
\cortext[mycorrespondingauthor]{Corresponding author}
\ead{pablo.del_moral@hh.se}

\author[mymainaddress]{S\l awomir Nowaczyk}
\author[mymainaddress]{Anita~Sant'Anna}
\author[mymainaddress]{Sepideh Pashami}

\address[mymainaddress]{CAISR, Halmstad University, Kristian IV:s väg 3, 301 18 Halmstad, Sweden}

\begin{abstract}
Using hierarchies of classes is one of the standard methods to solve multi-class classification problems. In the literature, selecting the right hierarchy is considered to play a key role in improving classification performance. Although different methods have been proposed, there is still a lack of understanding of what makes one method to extract hierarchies perform better or worse.

To this effect, we analyze and compare some of the most popular approaches to extracting hierarchies. We identify some common pitfalls that may lead practitioners to make misleading conclusions about their methods.
In addition, to address some of these problems, we demonstrate that using random hierarchies is an appropriate benchmark to assess how the hierarchy's quality affects the classification performance.

In particular, we show how the hierarchy's quality can become irrelevant depending on the experimental setup: when using powerful enough classifiers, the final performance is not affected by the quality of the hierarchy. We also show how comparing the effect of the hierarchies against non-hierarchical approaches might incorrectly indicate their superiority.

Our results confirm that datasets with a high number of classes generally present complex structures in how these classes relate to each other. In these datasets, the right hierarchy can dramatically improve classification performance.

\end{abstract}


\end{frontmatter}

%\linenumbers

\section{Introduction}
\label{intro}
Extracting hierarchies of classes to transform Multi-class Classification (MC) problems into Hierarchical Multi-class Classification (HMC) ones is a popular approach in the literature \cite{aly2005}. Compared to other approaches, it reduces training and especially testing running times while being competitive in terms of classification performance \citep{Bengio2010}. However, quite surprisingly, a thorough discussion on the nature and importance of the extracted hierarchies and how to extract and evaluate them is still missing.

\begin{figure}
	\centering
	\includegraphics[width=0.7\textwidth]{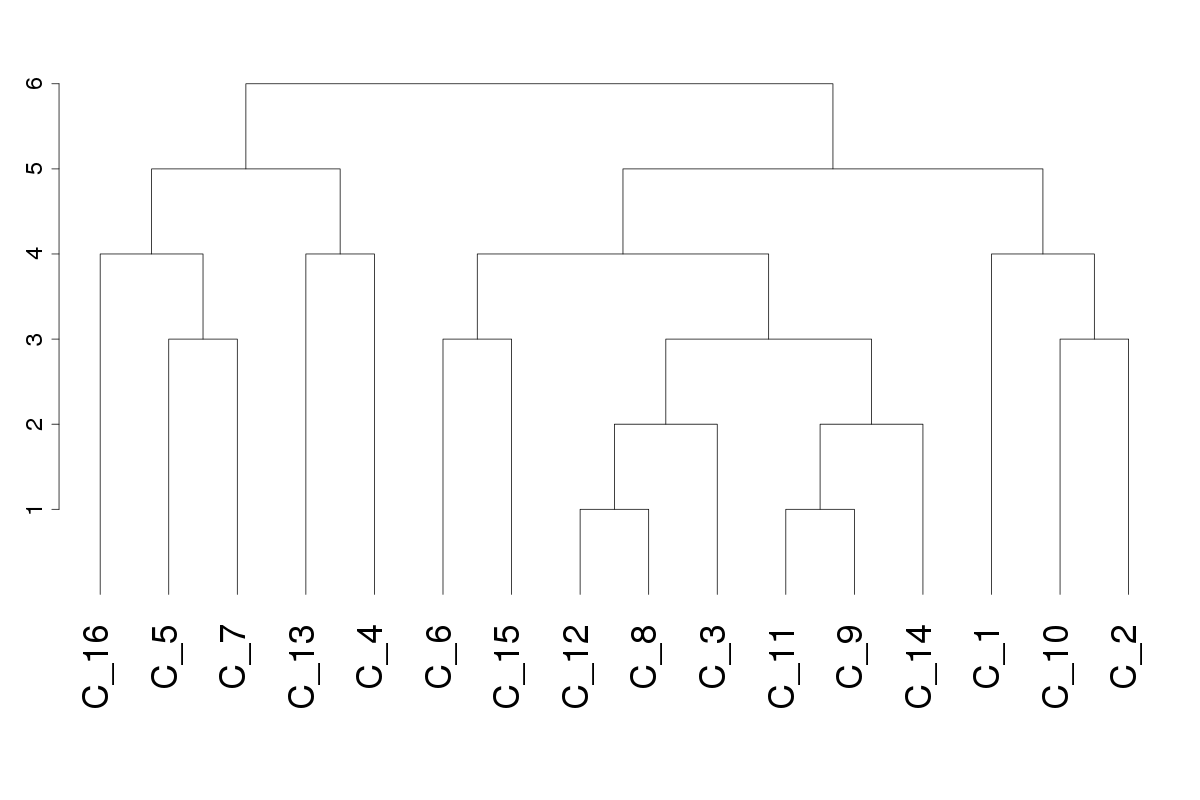}
	\caption{Example of a hierarchy of Classes}
	\label{fig:hierarchy}      
\end{figure}

Intuitively, HMC consists of breaking the multi-class problem into a collection of binary classification problems defined by a particular hierarchy. The hierarchy is a binary dendrogram, where all classes are placed at the leaves and are iteratively merged until they are all clustered together at the root (see Figure \ref{fig:hierarchy} for an example). At each internal node, a binary classifier is trained to discriminate the data in the left sub-tree from the data in the right sub-tree. When testing a new instance, the model starts by evaluating the classifier at the root node. In the subsequent steps, only the classifier at the node selected by the previous one is evaluated. This continues until a leaf is reached. Therefore, for a given instance, the total number of classifiers to be evaluated is equal to the number of nodes in the path from the final predicted class to the root and is always smaller than the total number of classes.

Hierarchical multi-class classification might suffer from the so-called "error propagation". If an instance is incorrectly classified at the early nodes, the error will propagate downwards. Since we are dealing with hard-class predictions, these errors will be unrecoverable. 

In this sense, the intuition behind a good hierarchy for classification is that easy to separate classes are clustered close to the root, while hard to separate ones are clustered together farther from the root and closer to the leaves. This way, the classifiers trained at the higher nodes of the hierarchy need to solve easy decision boundaries, which are less prone to error. On the other hand, the hard decision boundaries are isolated at the bottom of the dendrogram and tackled by a classifier trained explicitly for them.

Clearly, this idea defines the first assumption, or requirement, for the existence of a good hierarchy: that classes are not independent of each other and that there is some structure between them. In a sense, we assume that class {\it A} is closer to class {\it B} than to class {\it C} \citep{Rifkin_2004}. The next assumption is that hierarchies that capture these relationships will lead to a significant increase in the classification performance when compared to a hierarchy that does not incorporate this knowledge.

The total number of possible hierarchies is $(2n-3)!!$ where $n$ is the number of classes \citep{stanley1986}. Finding the best hierarchy through an optimization scheme will probably be costly, especially as the number of classes increases. However, if we assume that the best hierarchy is the one that best captures relationships between classes, we can focus on the process of extracting the hierarchy directly from the data in an unsupervised manner. Formulating the extraction of a hierarchy of classes as an unsupervised learning task will also help us find the hierarchy that best generalizes to unseen data.

Extracting hierarchies from the data for HMC has been heavily studied in the machine learning literature. HMC has been proposed independently in several works \citep{Vural2004, Silva-palacios2018, Li2007, Godbole2002}. Typically, the main approach was presented together with a methodology to extract the hierarchy. The overall performance was then compared with other approaches that do not involve hierarchies, such as \textit{single multi-class classifiers} or One-vs-All ensembles. 

This type of analysis fails to answer the following questions: is there a structure in how classes relate to each other? Is this method successfully discovering this structure? Is the extracted hierarchy useful for the learning algorithm used? In other words, these works validate the goodness of the hierarchical classification approach, but not necessarily the quality of the method to extract the hierarchy and the quality of the hierarchy itself.

In this paper, we answer these questions by comparing different methods of extracting hierarchies. Conceptually, extracting hierarchies from data points is a well-established field of research called hierarchical clustering. Any method in this area consists of two main components: measuring the dissimilarity between points and evaluating the quality of the hierarchy. In the specific context of HMC, extracting hierarchies presents two specific challenges. First, instead of measuring dissimilarities between points, one must measure dissimilarities between classes -- which can be thought of as sets of data points, but that is not necessarily the best approach since it ignores important properties such as the shape or complexity of class boundaries. Second, the quality of the hierarchy is characterized in an external way, namely by the performance of the classifier trained on it, instead of the inherent structure of the data.

We are going to critically evaluate different state-of-the-art approaches for obtaining hierarchies of classes, focusing on how to measure dissimilarity between classes and how to cluster them. In particular, we are going to compare two families of dissimilarities: Representative Based (RBD) and Classifier Based (CBD). RBD computes a representative for each class and measures dissimilarity between class representatives. On the other hand, CBD first trains a classifier on the whole dataset and infers dissimilarity from the resulting confusion matrix.  In addition, we are going to compare two examples of clustering algorithms: Hierarchical Agglomerative Clustering (HAC) and Hierarchical K-means (HKM). The first one represents the family of bottom-up algorithms, whereas the latter represents the family of top-down. We demonstrate that the quality of the hierarchy is not the only factor that affects the final classification performance, and identify several factors that must be taken into account when presenting evaluations. In particular, we analyze the effects of the HMC approach itself and different levels of usefulness of the hierarchy regarding how powerful the trained learning algorithms are.

This paper contributes to the state-of-the-art with the following:

\begin{itemize}\vspace{-0.5em}
	\item we identify several pitfalls in the process of extracting and evaluating methods to extract hierarchies in the context of HMC;
	\item we propose using a random hierarchy as a necessary benchmark to evaluate the relevance of a good hierarchy, in this way we can establish whether our extracted hierarchy is capturing an existing structure between classes or not;
	\item we demonstrate how the relevance of the hierarchy quality depends on the complexity of the classification problem and the complexity of the classifiers used;
	\item we perform experimental evaluation of different approaches for extracting hierarchies to showcase the previous contributions.
	
\end{itemize}

\section{Literature Review}
\label{sec:literature}

Traditionally, learning algorithms are designed to deal with binary classification problems. In this sense, multi-class classification approaches can be grouped into two categories: extended algorithms and binary adaptation. 

Extended algorithms refer to those algorithms that are adapted for the multi-class case by solving the multi-class problem as a global optimization problem, overlooking relationships between classes. In this paper, we will refer to them as single multi-class classifier, or simply \textit{single classifier}.

Some learning algorithms are easily extended, like KNN, decision trees, boosting algorithms, or neural networks. Others, like SVMs (\citep{weston1998}; \citep{lee2004}) or logistic regressions \citep{engel1988}, cannot be adapted so straightforwardly. In any case, the number of possible classes affects at least linearly the computational complexity of training and testing\footnote{KNN complexity is not affected by the number of classes, but it is quadratic in the number of instances, which makes it unsuitable for big problems.}.

Binary adaptation consists of decomposing the multi-class problem into a collection of binary classifiers. Depending on how the decomposition is done, we can further distinguish between approaches that take into account the structures and the relationships between classes and those that do not.

All-vs-all (AVA) and One-vs-all (OVA) are the two most popular approaches among the latter. AVA \citep{hastie1998} consists of training one classifier for each pair of classes. $\frac{N(N-1)}{2}$ classifiers must be trained. The final prediction is obtained after evaluating all classifiers through some form of ensemble aggregation method \citep{galar2011}. OVA \citep{Rifkin_2004} consists of training one classifier for each class, separating between the given class and the rest. The final prediction is obtained via voting as in the case of AVA \citep{galar2011}. 

In \citep{Rifkin_2004}, the authors conclude that there are no significant differences in performance between these OVA and AVA when the binary learners are well-tuned regularized classifiers. Moreover, they reflect on the limitations of OVA in the cases where classes are not independent, i.e., ``we do not necessarily expect examples from class {\it A} to be closer to those in class {\it B} than those in class {\it C}.'' They hypothesize that an approach that exploits these relationships can achieve better performance. This hypothesis was separately corroborated from a theoretical point of view in \citep{liu2017} and \citep{Babbar2013}.

The two main approaches that focus on exploiting the relationship between classes to improve performance are Chains of Classifiers (CC) and Hierarchical Multi-class Classification (HMC). CC \citep{liu2017} consists of training a sequence of binary classifiers. The first classifier distinguishes one class from the rest; the next one separates another single class from the remaining rest. At each step, the output of all the previous classifiers is fed to the training set. The process is repeated until all classes are classified. Dissimilarity between classes is used to establish the order of the classes so that easy to discern classes are classified first and hard to discern classes are classified at the end of the chain.

\subsection{Hierarchical Multi-class Classification}

Hierarchical Multi-class Classification is based on using a hierarchy of classes using a dendrogram structure. Each class corresponds to one leaf of the tree. At each node, a classifier is trained to distinguish among its children. Predictions start from the root node and are propagated until a leaf is reached. Contrary to the CC method, the outputs of the parent classifier is not used for training their children.

Testing time is reduced, compared to AVA or OVA \citep{Vural2004}, where each node produces probabilistic predictions and these propagate through the tree, i.e., all nodes need to be evaluated. This benefit is significantly increased when hard predictions are propagated. In this setup, only the nodes connecting the final predicted class and the root need to be evaluated \citep{Bengio2010}. Reduction of testing time comes at the expense of reducing the performance. In \citep{Qu2017}, the authors explore this trade-off by allowing the evaluation of the most probable paths, instead of just one.

The term "hierarchical classification" has been used both for tasks where there is a pre-existing hierarchy of classes that can be exploited \citep{silla2011}, or for tasks where such hierarchy does not exist and needs to be extracted \citep{Chen2004}. Even in the cases where there is a predefined hierarchy, improvements can be achieved by modifying it. In \citep{babbar2016}, hierarchies are modified by eliminating some of the intermediate nodes of the hierarchy. In \citep{naik2019}, predefined hierarchies can be modified by rewiring the hierarchy changing parent-child relations or deleting and creating intermediate nodes. We will focus on the extraction of hierarchies for multi-class classification problems where there is not a predefined one. 

Extraction of hierarchies has also been used in the field of neural networks and language modeling. In cases with many possible outputs, hierarchical softmax provides significant speed-up for both training and testing \citep{morin2005}. In this work, a hierarchy of classes is provided and modified to improve classification performance. Extraction of hierarchies of words was explored in \citep{mnih2009}, yielding significant improvements in performance.

Hierarchical Multi-Class classification with binary splitting is often referred to as nested dichotomies. In \citep{Frank_2004}, they achieve improvements in predictive power by ensembling several randomly generated hierarchies. In \citep{melnikov2018}, they study the effects of random hierarchies and relate them with the learning algorithm used for classification. An entirely new approach is presented to obtain hierarchies based on nested dichotomies: instead of extracting the hierarchy from the data, they choose the best hierarchy from a pool of randomly generated ones. 

In this paper, we will focus on extracted hierarchies with just binary splittings where only hard class propagation is considered, and the effect they have in this setup. We choose to keep the term "Hierarchical Classification", because we believe that our contributions can be easily generalized to less constrained setups.

Focusing on the process of extracting the hierarchy, we distinguish two steps: selecting the dissimilarity measure to be used and a hierarchical clustering algorithm. 

Dissimilarity measures between classes can be divided into two groups: Representative Based Dissimilarity (RBD) and Classifier Based Dissimilarity (CBD). RBD consists of obtaining a representative (or a prototype) for each class and measuring distances between those representatives. In (\citep{Larsen_1999}; \citep{Steinbach00}; \citep{Li2007}) authors select a centroid for each class and measure the cosine dissimilarity in a problem of document classification. A similar approach is used in \citep{Vural2004}, except for the usage of the euclidean distance between centroids. The main advantage of using this type of dissimilarity is computational efficiency. In fact, in \citep{Vural2004}, they propose a clustering algorithm based on a random split that respects the balance of the clusters so that running time is further reduced.

While RBD is simple and fast, it also presents some drawbacks. Namely, how representative is the centroid of a class and how the distances between centroids describe the relations between classes. To answer these drawbacks, CBD builds on the concept of separability. The intuition behind using CBD is that if two classes are frequently misclassified into each other, they will be similar, whereas if the classes are not mixed, they are dissimilar.

CBD involves training an auxiliary classifier and evaluating the confusion matrix between classes. To the best of our knowledge, the first contribution among these approaches can be found in \citep{Godbole2002}. The author calculates the confusion matrix using a single classifier, then uses the rows of the matrix as class representative for the corresponding class. In \citep{Bengio2010}, the confusion matrix is calculated using the OVA approach as a classification method; then the confusion matrix is symmetrized and fed to a spectral clustering algorithm. In \citep{Silva-palacios2018}, a single classifier is trained, then, the dissimilarity between two classes is calculated based on the number of instances of each class wrongly classified as the other.

Once dissimilarities have been computed, a hierarchical clustering algorithm must be chosen. These algorithms can be divided into agglomerative and divisive. Hierarchical Agglomerative Clustering (HAC) consists of progressively merging the most similar objects and is probably the most basic of these methods. It is used in \citep{Silva-palacios2018}, and \citep{Godbole2002}. Divisive clustering algorithm consists of iteratively using flat cluster algorithms until each class is separated from the rest. K-means is used in \citep{Larsen_1999} and \citep{Vural2004}, whereas \citep{Bengio2010}, and \citep{Chennupati2016} use spectral clustering algorithms. Other divisive methods aim to create balanced hierarchies in order to improve training and testing complexities (cf.  \citep{Tsoumakas2008}; \citep{Vural2004}).

Hierarchical Multi-Class classification with binary splitting is often referred to as nested dichotomies. In \citep{Frank_2004}, they achieve improvements in predictive power by ensembling several randomly generated hierarchies. In \citep{melnikov2018}, they study the effects of random hierarchies and relate them with the learning algorithm used for classification. An entirely new approach is presented to obtain hierarchies based on nested dichotomies: instead of extracting the hierarchy from the data, they choose the best hierarchy from a pool of randomly generated ones.

\section{Methods}

This section provides a structured overview of the methods that we use for extracting class hierarchies for multi-class classification.

Formally, a hierarchy is defined as a binary dendrogram. Starting from the root, all classes are iteratively separated into two sub-trees until each class is placed in its own leaf of the tree. An example of such a hierarchy can be seen in Figure \ref{fig:hierarchy}. Methods for extracting hierarchies from the data consist of two critical components: measuring the dissimilarity between classes and the hierarchical clustering algorithm.

\subsection{Dissimilarity}

A class is defined by a set of points annotated with the same label. Evaluating the dissimilarity between two classes $c_1, c_2$ means evaluating a function of the form
$d(c_1,c_2)=f(D_{c_1} , D_{c_2})$, where $D_{c}=\{\mathbf{x^i_c}\}$ is the distribution of points defined by the feature vectors $\mathbf{x^i}$ annotated with label $c$.

\paragraph{\textbf{Representative Based Distance (RBD)}}
One way to measure the dissimilarity between classes consists of obtaining a class representative and evaluating their distance. This approach is frequently used in the literature, using the centroid as the class representative. Once a single data point per class is chosen, any distance metric can be used to calculate the dissimilarity. It is a straightforward approach and usually quite favorable in terms of computational complexity. Formally, the dissimilarity between two classes becomes

\begin{equation}
    d(c_1, c_2) = d'(\overline{\mathbf{x}}_{c_1}, \overline{\mathbf{x}}_{c_2})
\end{equation}
% $$ d(c_1, c_2) = d'(\overline{\mathbf{x}}_{c_1}, \overline{\mathbf{x}}_{c_2}),$$

where $d'$ can be any distance metric (popular choices include euclidean or cosine distance).

\paragraph{\textbf{Classifier Based Dissimilarity (CBD)}}

The intuition behind using CBD is that if a classifier can not distinguish clearly between two classes, those classes must be similar. Conversely, if a classifier can clearly differentiate between two classes, those two classes must be dissimilar. The most standard way of evaluating how well a classifier distinguishes between two classes is by evaluating its accuracy. 

For a multi-class classification problem, evaluating all pairwise class dissimilarities is similar to using the All-vs-all approach. If the number of classes is high, this task can become extremely time-consuming. Practitioners avoid this problem by obtaining a confusion matrix using faster approaches like OVA or single classifier. One option is to use the confusion matrix's rows as the corresponding class representatives \citep{Godbole2002, Bengio2010}. Another approach \citep{Silva-palacios2018} is to calculate the dissimilarity between two classes as the accuracy measured in the corresponding subset of the confusion matrix $M$. 

\begin{align}
    M=
\begin{bmatrix}
m_{11}       & m_{12} & m_{13} \\
m_{21}       & m_{22} & m_{23} \\
m_{31}       & m_{32} & m_{33}
\end{bmatrix}
% =>
% \begin{bmatrix}
% m_{11} & m_{13}  \\
% m_{31} & m_{33} \\
% \end{bmatrix}
=> d(c_1, c_3) = \frac{m_{11}+m_{33}}{m_{13}+m_{31}+m_{11}+m_{33}}
\end{align}
% \[

% \]

As opposed to the AVA scenario, these approaches not only evaluate how two classes relate to each other but also take into account the relation to the rest of the classes. The relationship between two classes can be overshadowed by the overlap with a third one.

\paragraph{\textbf{All-vs-all testing.}}
\label{subsec:avatesting}

To avoid this problem and still use the faster OVA approach, we propose to use the output of a probabilistic classifier as a proxy to evaluate the accuracy of distinguishing between two classes in the AVA fashion. 

Let's consider a test set $D=[(\mathbf{x^1},y^1)... (\mathbf{x^m},y^m)] $ where $\mathbf{x^i}$ is the feature vector,  $y^i \in \{c_1, ... c_n\}$ is the associated class. Let $\mathbf{h(x)} = [P(c_1|\mathbf{x}), \dots, P(c_n|\mathbf{x})]$ be the output of our probabilistic multi-class classifier. For each pair of classes $c_j$ and $c_k$, we evaluate the classifier only on those instances with real labels $c_j$ and $c_k$ and only compare the probability predictions corresponding to these classes. We define the proxy of the binary classifier for classes $c_j$ and $c_k$ as:

\begin{equation}
    h'(\mathbf{x}) = \argmax_{c_j,c_k} (P(c_j|\mathbf{x}), P(c_k|\mathbf{x}))
\end{equation}
% $$ h'(\mathbf{x}) = \argmax_{c_j,c_k} (P(c_j|\mathbf{x}), P(c_k|\mathbf{x})) $$

Finally, the distance between classes $c_j$ and $c_k$ is:
\begin{equation}
d(c_j, c_k) = \frac{\sum\limits_{i: y_i \in \{c_j, c_k \}} (h'(\mathbf{x_i})=y_i) }{\sum_i (y_i= c_j)+ \sum_i (y_i= c_k)}
\end{equation}

\subsection{Building the hierarchy}

\paragraph{\textbf{Hierarchical Agglomerative Clustering (HAC)}} is an algorithm that works in a bottom-up fashion. Initially, the data from each class becomes its own cluster. Then, given a set of clusters, their dissimilarities can be calculated in different ways. The most popular is the ``Average Link'' approach, where the distance between two clusters is the average distance between all pairs of points. HAC then merges the two most similar clusters, creating a new intermediate node in the hierarchy. The process continues until all of the data is combined in the root node.

\paragraph{\textbf{Hierarchical K-means}} is an example of a divisive hierarchical clustering algorithm. It works in a top-down manner by iteratively separating the points into two groups. Starting from the root, we use k-means with $k=2$ to create two sub-clusters of classes. We recursively repeat this step on each sub-cluster until every class is separated from the rest and placed at one leaf. 

\begin{table}[!t]
	\renewcommand{\arraystretch}{1.3}
	
	\centering
	\begin{tabular}{lc|ccccc}
		\hline
		&Dataset name 	& Instances		& Attributes 	& Classes & Min & Max\\
		\hline
		1&Collins & 1000 & 19 & 30 & 6 & 80\\
		2&Pen digits & 10992 & 16 & 10 & 1055 &1144\\
		3&Abalone & 4168 & 9 & 21 &  6 &  689\\
		4&Arrythmia & 438 & 262 & 9 & 9 &245\\
		5&Image Segmentation & 2310 & 18 & 7 & 330 &330\\
		6&Cartdiotocography & 2126 & 35 & 10 & 53 &579\\
		7&Semeion & 1593 & 256 & 10 & 155 &162\\
		8&Texture & 5500 & 40 & 11 & 500 &500\\
		9&Statlog(Sat) & 6430 & 36 & 6 & 625 &1531\\
		10&Walking & 149332 & 4 & 22 & 911 &21991\\
		11&Usps & 9298 & 256 & 10 & 708 &1553\\
		12&Bach & 5571 & 90 & 65 & 6 &503\\
		13&Letters & 20000 & 16 & 26 & 734 &813\\
		14&Helena & 65196 & 27 & 100 & 111 &4005\\
		15&Plants(margin) & 1600 & 64 & 100 & 16 &16\\
		16&Plants(shape) & 1600 & 64 & 100 & 16 &16\\
		17&Plants(texture) & 1599 & 64 & 100 & 15 &16\\
		18&Mnist & 7000 & 719 & 10 & 6313 &7877\\
		19&Covertype & 581012 & 54 & 7 & 2747 &283301\\
		\hline
	\end{tabular}
	\caption{Information about dataset used: number of instances, number of features, number of classes, and number of instances for the most and lest frequent class.}
	\label{table_1}
\end{table}

\section{Experiments and Results}
\label{sec:results}

All experimental results reported are the average accuracy measured over 20 fold Monte Carlo validation using 90\% of the data for training and 10\% for testing each fold. Variances in the accuracies are computed with the method presented in \citep{Nadeau2003}. Hypothesis testing is computed using the corrected resampled t-test also presented in \citep{Nadeau2003}.  

Each iteration consists of measuring the dissimilarities, extracting the hierarchy with the corresponding hierarchical clustering algorithm, training binary classifiers on each node of the hierarchy, and evaluating the results. Representative Based Dissimilarities (RBD) are calculated directly, since it is a deterministic measure, while Classifier Based Dissimilarities (CBD) are estimated as an average over 10 fold Monte Carlo validation. For clarity, we will refer to the classifiers trained to measure CBD dissimilarities as \textbf{CBD classifiers}; we will refer to the classifiers trained on the hierarchy as \textbf{base classifiers}. The \textit{base classifiers} trained at each node of the hierarchy use 3 fold cross-validation to tune the parameters if needed. 

To sample hierarchies uniformly, we follow the method proposed in \citep{melnikov2018}.
The datasets used are publicly available in the UCI and Open ML repositories\footnote{\url{https://archive.ics.uci.edu/ml/datasets.html}, \url{https://www.opneml.org}} and an overview is presented in Table \ref{table_1}.

\subsection{Benchmarks vs random hierarchy}
\label{sec:random hierarchy}

\begin{table}[t]
	\centering
	\begin{tabular}{l|c|c|c|}
		
		&Single Classifier& Random  & OVA \\
		\hline
		1&\underline{0.07805$\rpm$0.0085}$^{\downarrow\downarrow\downarrow}$	&0.142$\rpm$0.015		&0.166$\rpm$0.013\\
		2&\underline{0.902$\rpm$0.017}$^{\downarrow\downarrow}$		&0.9484$\rpm$0.0032		&0.9487$\rpm$0.0024\\
		3&0.254$\rpm$0.0064		&0.2417$\rpm$0.0076		&0.2462$\rpm$0.0094\\
		4&\underline{0.6$\rpm$0}$^{\downarrow\downarrow\downarrow}$			&0.738$\rpm$0.019		&0.718$\rpm$0.027\\
		5&\underline{0.926$\rpm$0.0068}$^{\downarrow\downarrow}$		&0.9556$\rpm$0.0042		&0.9576$\rpm$0.0053\\
		6&0.99928$\rpm$0.00069	&0.9988$\rpm$0.00084	&1$\rpm$0\\
		7&\underline{0.523$\rpm$0.018}$^{\downarrow\downarrow\downarrow}$		&0.74$\rpm$0.016		&0.762$\rpm$0.013\\
		8&\underline{0.8433$\rpm$0.009}$^{\downarrow\downarrow\downarrow}$		&\textbf{0.9245$\rpm$0.004}$^{\uparrow\uparrow}$		&0.9093$\rpm$0.0042\\
		9&\underline{0.8383$\rpm$0.0077}$^{\downarrow}$	&\textbf{0.8627$\rpm$0.0041}$^{\uparrow}$		&0.8546$\rpm$0.0048\\
		10&\underline{0.5099$\rpm$0.0016}$^{\downarrow\downarrow\downarrow}$	&0.5986$\rpm$0.002		&\textbf{0.6209$\rpm$0.0017}$^{\uparrow\uparrow\uparrow}$\\
		11&\underline{0.8059$\rpm$0.0039}$^{\downarrow\downarrow\downarrow}$	&0.8565$\rpm$0.0091		&\textbf{0.8758$\rpm$0.0056}$^{\uparrow}$\\
		12&\underline{0.6643$\rpm$0.0064}$^{\downarrow\downarrow\downarrow}$	&0.7236$\rpm$0.0044		&0.7269$\rpm$0.0065\\
		13&\underline{0.7208$\rpm$0.0043}$^{\downarrow\downarrow\downarrow}$	&0.8541$\rpm$0.0023		&0.8547$\rpm$0.0028\\
		14&\underline{0.2083$\rpm$0.0011}$^{\downarrow\downarrow\downarrow}$	&0.2368$\rpm$0.0049		&\textbf{0.2859$\rpm$0.0022}$^{\uparrow\uparrow\uparrow}$\\
		15&\underline{0.021$\rpm$0.0017}$^{\downarrow\downarrow\downarrow}$	&0.348$\rpm$0.018		&0.355$\rpm$0.012\\
		16&\underline{0.02$\rpm$0}$^{\downarrow\downarrow\downarrow}$			&0.309$\rpm$0.018		&0.322$\rpm$0.016\\
		17&\underline{0.01$\rpm$0}$^{\downarrow\downarrow\downarrow}$			&0.398$\rpm$0.014		&\textbf{0.441$\rpm$0.013}$^{\uparrow}$\\
		18&\underline{0.88098$\rpm$0.00059}$^{\downarrow\downarrow\downarrow}$	&0.9356$\rpm$0.0011		&\textbf{0.9394$\rpm$0.00052}$^{\uparrow\uparrow}$\\
		19&\underline{0.8184$\rpm$0.0039}$^{\downarrow\downarrow\downarrow}$	&0.8597$\rpm$0.0057		&\textbf{0.8811$\rpm$0.0018}$^{\uparrow\uparrow}$\\
		\hline
	\end{tabular}
	
	\caption{Comparison of single classifier, Hierarchical Multi-class Classification using a random hierarchy, and One-vs-All (using CART classifier). Highlighted are those results that are significantly better than the other two options and underlined are those significantly worse than the other two. One arrow indicates a p-value of 0.05; two arrows, 0.01; three arrows, 0.001.}
	\label{table:random_comparison}
\end{table}

To evaluate the quality of the extracted hierarchy, one needs an appropriate benchmark to compare against. If we compare Hierarchical Multi-class Classification (HMC) against other approaches like One-vs-All (OVA) or a single classifier, we are evaluating the quality of HMC as an approach and the quality of the hierarchy simultaneously. We claim that comparing against a random hierarchy is a suitable solution to decouple the effects of using HMC and isolate the effects of the hierarchy's quality. 

In this experiment, we are comparing the results of a classifier trained on a random hierarchy with the two popular benchmarks used in the literature: single classifier and OVA. Our goal is to understand how the characteristics of HMC affects performance.

As the learning algorithm within the nodes of the hierarchy, we use CART (as implemented in the \textbf{rpart} R package). We select the best complexity parameter from a grid going from $10^0$ to $10^{-6}$. 

In Table \ref{table:random_comparison}, we present the results of the experiment. The performance of the single classifier is significantly worse than HMC on 17 out of 19 datasets. In the remaining two cases, there is no significant difference between the two approaches. Clearly, this demonstrates that a single classifier cannot be used as a benchmark for evaluating the quality of the hierarchy; even HMC trained on a random hierarchy generally outperforms it.

On the other hand, OVA produces significantly better results in 6 out of 19 datasets than HMC trained on a random hierarchy. However, in 2 cases, HMC trained on a random hierarchy presents significantly better results, and for the remaining datasets, there is no significant difference between these two approaches. OVA is not expected to outperform HMC, even when trained on a random hierarchy, which makes OVA an inappropriate benchmark to evaluate the quality of the extracted hierarchies.

With this experiment, we have proven that Hierarchical Multi-class Classification has benefits independent of the quality of the hierarchy chosen. Using randomly generated hierarchies can result in performances significantly better than non-hierarchical approaches. If we want to evaluate methods of extracting hierarchies, comparing against methods that do not involve hierarchies gives an incomplete evaluation. Nevertheless, this type of analysis is still used in, for example, \citep{Vural2004}, \citep{Li2007}, \citep{Godbole2002}, \citep{morin2005}, \citep{Chennupati2016}.

In \citep{melnikov2018}, the authors propose a method to evaluate the quality of the hierarchy by comparing it against random hierarchies. The exceedance probability ranks the performance of a model trained on a given hierarchy against a number of random hierarchies (in their case, 10000). Training and testing 10000 hierarchies is costly. Also, there is the question of how many random hierarchies are needed to give a fair representation of all possible hierarchies. For example, for a problem with 100 classes, there are approximately $10^{184}$ possible hierarchies.

In the rest of the paper, we will compare different approaches to extract hierarchies. Random hierarchies must be part of the process of evaluating the quality of the extracted hierarchies. Instead of the exceedance probability, we will compare against the performance of a hierarchical classifier trained on a hierarchy randomly sampled. This comparison does not allow us to determine how good a hierarchy is. However, it allows us to decouple the effects of the hierarchy's quality from the effects on the performance inherent in HMC approaches. Suppose a given hierarchy obtains significantly better classification performance when compared with a random hierarchy. In that case, we can conclude that this increase is due to the quality of the hierarchy and that this is capturing some relationship between the classes.

\subsection{Random hierarchies vs informed hierarchies}
\label{sec:random_vs_informed}

Random hierarchies do not incorporate any information about the relationships between classes. The main hypothesis of HMC is that an informed hierarchy that exploits such information should yield better results in terms of predictive accuracy. However, we demonstrate that this is not necessarily universally true. We hypothesize that how relevant good hierarchies are actually depends on the learning algorithm used as the base classifier. With powerful classifiers, the hierarchy chosen will have a smaller effect than it would with a weaker classifier.

\begin{table}[h]
	\centering
	\begin{tabular}{l|c|c||c|c|}
		& \multicolumn{2}{c||}{CART} & \multicolumn{2}{c|}{Logistic}\\
		\cline{2-5}
		&Random  & RBD & Random  & RBD   \\
		\hline
		1&0.142$\rpm$0.015&0.171$\rpm$0.016&            0.195$\rpm$0.012&\textbf{0.248$\rpm$0.016}$^{\uparrow\uparrow}$\\
		2&0.9484$\rpm$0.0032&\textbf{0.9558$\rpm$0.0024}$^{\uparrow}$&        0.86$\rpm$0.016&\textbf{0.9184$\rpm$0.0027}$^{\uparrow\uparrow}$\\
		3&0.2417$\rpm$0.0076&0.229$\rpm$0.011&          0.225$\rpm$0.011&0.2448$\rpm$0.007\\
		4&0.738$\rpm$0.019&0.767$\rpm$0.019&            0.509$\rpm$0.028&0.507$\rpm$0.034\\
		5&0.9556$\rpm$0.0042&0.9606$\rpm$0.0058&        0.888$\rpm$0.02&\textbf{0.9429$\rpm$0.0056}$^{\uparrow\uparrow}$\\
		6&0.9988$\rpm$0.00084&0.99952$\rpm$0.00058      &1$\rpm$0&1$\rpm$0\\
		7&0.74$\rpm$0.016&\textbf{0.792$\rpm$0.015}$^{\uparrow\uparrow}$&             0.719$\rpm$0.023&\textbf{0.776$\rpm$0.011}$^{\uparrow}$\\
		8&0.9245$\rpm$0.004&0.9346$\rpm$0.0037&         0.9824$\rpm$0.0052&\textbf{0.99582$\rpm$0.00098}$^{\uparrow}$\\
		9&0.8627$\rpm$0.0041&0.8659$\rpm$0.0053&        0.8249$\rpm$0.0093&\textbf{0.8488$\rpm$0.0051}$^{\uparrow\uparrow}$\\
		10&0.5986$\rpm$0.002&0.6$\rpm$0.0018&           0.206$\rpm$0.012&0.2114$\rpm$0.0013\\
		11&0.8565$\rpm$0.0091&\textbf{0.8944$\rpm$0.0037}$^{\uparrow\uparrow\uparrow}$&       0.85$\rpm$0.016&\textbf{0.9084$\rpm$0.0039}$^{\uparrow\uparrow\uparrow}$\\
		12&0.7236$\rpm$0.0044&\textbf{0.747$\rpm$0.0085}$^{\uparrow\uparrow}$&        0.6679$\rpm$0.0077&\textbf{0.6949$\rpm$0.0076}$^{\uparrow\uparrow}$\\
		13&0.8541$\rpm$0.0023&\textbf{0.8628$\rpm$0.0036}$^{\uparrow}$&       0.483$\rpm$0.022&\textbf{0.7359$\rpm$0.0034}$^{\uparrow\uparrow\uparrow}$\\
		14&0.2368$\rpm$0.0049&\textbf{0.2688$\rpm$0.0029}$^{\uparrow\uparrow\uparrow}$&       0.2003$\rpm$0.0064&\textbf{0.2661$\rpm$0.0029}$^{\uparrow\uparrow\uparrow}$\\
		15&0.348$\rpm$0.018&\textbf{0.496$\rpm$0.017}$^{\uparrow\uparrow\uparrow}$&           0.26$\rpm$0.019&\textbf{0.371$\rpm$0.015}$^{\uparrow\uparrow\uparrow}$\\
		16&0.309$\rpm$0.018&\textbf{0.392$\rpm$0.015}$^{\uparrow\uparrow}$&           0.121$\rpm$0.012&\textbf{0.269$\rpm$0.018}$^{\uparrow\uparrow\uparrow}$\\
		17&0.398$\rpm$0.014&\textbf{0.517$\rpm$0.017}$^{\uparrow\uparrow\uparrow}$&           0.337$\rpm$0.02&0.361$\rpm$0.014\\
		18&0.9356$\rpm$0.0011&\textbf{0.93855$\rpm$0.00035}$^{\uparrow\uparrow}$&     0.7008$\rpm$0.0048&\textbf{0.7186$\rpm$0.0022}$^{\uparrow\uparrow}$\\
		19&0.8597$\rpm$0.0057&\textbf{0.8784$\rpm$0.0018}$^{\uparrow\uparrow}$&       0.79$\rpm$0.013&\textbf{0.8549$\rpm$0.0015}$^{\uparrow\uparrow\uparrow}$\\
		\hline
	\end{tabular}
	
	\caption{Comparison of Hierarchical Multi-class Classification trained on a random hierarchy and on a hierarchy using Representative Based Dissimilarity. CART and logistic regression are used as learning algorithm in the nodes. Highlighted are the results that are significantly better than the rest. One arrow indicates a p-value of 0.05; two arrows, 0.01; three arrows, 0.001.}
	\label{table:random_RBED_CART_logi}
\end{table}

In this experiment, we evaluate one of the most popular methods to extract hierarchies, \textit{Representative Based Dissimilarity}. RBD takes a representative for each class, computes their dissimilarity, and cluster them in a hierarchy. As a representative, we use the centroid; as dissimilarity, the euclidean distance; as clustering algorithm, \textit{Hierarchical Agglomerative Clustering}.

The overall experimental setup is the same as the previous experiment. The comparison uses two base learners: CART and logistic regression (GLM, as implemented in the R package \textbf{stats}). Both learning algorithms are trained on the same training set and on the same extracted hierarchy.

Results are presented in Table \ref{table:random_RBED_CART_logi}. For both CART and GLM, hierarchies obtained with RBD always yield similar or significantly better results than the random hierarchy. This indicates that the informed hierarchies have indeed captured some existing relationships between classes that are then exploited by the classifier.

For CART, when compared with the random hierarchy, we observe significant improvements in 10 out of 19 datasets for the RBD hierarchy. In 8 datasets, we can not observe significant differences between any of the approaches.

For logistic regression, when compared with the random hierarchy, we observe significant improvements in 14 out of 19 datasets for the RBD hierarchy. In 4 datasets, we can not observe significant differences between the approaches.

\begin{table}[h]
	\centering
	\begin{tabular}{l|c|c|}
		
		&Random & RBD \\
		\hline
		1&0.257$\rpm$0.016&0.267$\rpm$0.018\\
		2&0.9895$\rpm$0.0014&0.9908$\rpm$0.0011\\
		3&0.262$\rpm$0.011&0.253$\rpm$0.011\\
		4&0.762$\rpm$0.018&0.767$\rpm$0.017\\
		5&0.9784$\rpm$0.0031&0.9823$\rpm$0.0038\\
		6&0.99952$\rpm$0.00058&0.99976$\rpm$0.00042\\
		7&0.9224$\rpm$0.0067&0.9328$\rpm$0.0071\\
		8&0.9814$\rpm$0.0017&0.9835$\rpm$0.0017\\
		9&0.9186$\rpm$0.0029&0.9192$\rpm$0.0032\\
		10&0.6661$\rpm$0.0018&0.6664$\rpm$0.0017\\
		11&0.9603$\rpm$0.0024&\textbf{0.9653$\rpm$0.003}\\
		12&0.7844$\rpm$0.0064&0.7907$\rpm$0.0062\\
		13&0.9562$\rpm$0.0016&0.9568$\rpm$0.0018\\
		14&0.3582$\rpm$0.0017&\textbf{0.3673$\rpm$0.0017}\\
		15&0.64$\rpm$0.025&\textbf{0.716$\rpm$0.012}\\
		16&0.509$\rpm$0.019&0.544$\rpm$0.016\\
		17&0.668$\rpm$0.018&\textbf{0.715$\rpm$0.015}\\
		18&0.9683$\rpm$0.0021&\textbf{0.9704$\rpm$0.0017}\\
		19&0.9736$\rpm$0.001&\textbf{0.97547$\rpm$0.00074}\\
		\hline
	\end{tabular}
	
	\caption{Comparison of Hierarchical Multi-class Classification trained on a random hierarchy and on a hierarchy using Representative Based Dissimilarity. Extreme gradient boosting. Highlighted are the results that are significantly better than the rest. One arrow indicates a p-value of 0.05; two arrows, 0.01; three arrows, 0.001.}
	\label{table:random_RBED_xgb}
\end{table}

The first conclusion is that informed hierarchies are always better or as good as random hierarchies. When the difference in performance is significant, we can confirm that there is a structure between classes and that this method to extract hierarchies is capturing (at least partially) this structure. In the absence of significant differences, we can neither confirm nor refute that there exists a structure in how classes relate to each other. This can mean two different things: either there does not exist a good hierarchy (one that produces significant improvements), or we have not found it yet.

It is worth noting that if we had compared with the single classifiers from the previous subsection, we would have concluded that the RBD hierarchies for datasets 1,  4, 5, 9, and 10 were useful. This conclusion would be wrong: the results of HMC trained on them are significantly better than the single classifier; however, they are not significantly better than HMC trained on a random hierarchy.

In all datasets where the differences were significant for CART, we have also found significant differences using GLM as a base classifier. However, there are multiple examples where the informed hierarchy has significantly outperformed random hierarchy using logistic regression, but the two hierarchies were equivalent from CART's perspective.

This suggests that powerful enough classifiers will not be affected by the quality of the hierarchy at all. In general, though, these powerful classifiers will be much more expensive to train and deploy. CART, being a more powerful classifier, is less sensitive to the quality of the hierarchies. The analysis of the results using logistic regression suggests that RBD hierarchies are indeed capturing some inherent relationships between classes. The absence of a significant difference using CART does not necessarily mean that the extracted hierarchy is not capturing interesting relationships between classes. Rather, CART can discover these relationships by itself.

As a natural extension, we hypothesize that for an arbitrarily complex problem, there is an arbitrarily complex learning algorithm for which the different possible decompositions will be irrelevant. This makes the question of extracting, in the abstract, ``the best'' hierarchy quite moot -- the quality of a hierarchy is always dependent on the context and its intended use.

In Table, \ref{table:random_RBED_xgb}, we can see the results of running a boosting algorithm based on trees -- an even more powerful classifier. We have used the \textbf{xgboost} package, and we have simply tuned the depth of the trees. Although we still find significant differences in some datasets, those differences are comparatively smaller and less significant when using CART or logistic regression. This result aligns with the findings of \citep{Rifkin_2004}: with a powerful enough classifier, properly tuned, the way to decompose the original problem does not matter. In our experiments, we have datasets with different properties, some of which are arguably simpler than others. 

If the practitioner is interested in finding the best possible classification performance without regard for the time it may take to train and evaluate the models; the effort should be directed into finding the most powerful classifier and tuning its hyperparameters. In these cases, there is probably no added value from choosing one decomposition over another. It is interesting to find good hierarchies when there are restrictions on how much resources we can afford for the training and evaluation phase.

At each fold of our validation scheme, we extract a hierarchy and train our models. In this scenario, our results' source of variation is not just coming from the sampling of training and testing sets, like in traditional classification tasks. There is an added contribution to the variance arising from the method chosen to extract hierarchies. The latter depends not only on how we design our validation scheme but also on the intrinsic complexity of the relationship between classes and how we try to code them in a hierarchy. This effect is evident by comparing the variances using a random hierarchy or an RBD hierarchy using logistic regression as a base classifier in Table \ref{table:random_RBED_CART_logi}.

Therefore, to establish a fair comparison between two methods to extract hierarchies over several datasets, we need first to make sure that there is an underlying structure that provides statistically significant improvements for each dataset. First, we need to analyze if the proposed method provides significant improvements over a random hierarchy, then we need to provide a sound statistical analysis in each dataset. This type of analysis is missing in \citep{Silva-palacios2018} and \citep{melnikov2018}.

\subsection{Representative Based Dissimilarity vs Best of 50 heuristic}
\label{sec:RBD vs Bo50}

RBD using centroid as class representative and euclidean distance as metric is a standard procedure to measure dissimilarities between classes. There are some concerns, though: how representative is the centroid for the complete class distribution (which can be arbitrarily complex) and how useful is the distance between centers to measure dissimilarity between classes. 

Best-of-50 (Bo50) method was introduced in \citep{melnikov2018}, it samples 50 hierarchies at random and selects the one with the best performance using a 3-fold cross-validation criteria.

\begin{table}[h]
	\centering
	\begin{tabular}{l|c|c||c|c|}
		& \multicolumn{2}{c||}{CART} & \multicolumn{2}{c|}{Logistic}\\
		\cline{2-5}
		&Bo50  & RBD & Bo50  & RBD   \\
		\hline
		1&0.185$\rpm$0.013&0.171$\rpm$0.016&            0.276$\rpm$0.015&0.248$\rpm$0.016\\
		2&0.9599$\rpm$0.0021&0.9558$\rpm$0.0024&        \textbf{0.9347$\rpm$0.0044}$^{\uparrow\uparrow\uparrow}$&0.9184$\rpm$0.0027\\
		3&\textbf{0.269$\rpm$0.0057}$^{\uparrow\uparrow\uparrow}$&0.229$\rpm$0.011&           \textbf{0.2726$\rpm$0.0057}$^{\uparrow\uparrow}$&0.2448$\rpm$0.007\\
		4&0.769$\rpm$0.018&0.767$\rpm$0.019&            \textbf{0.576$\rpm$0.031}$^{\uparrow}$&0.507$\rpm$0.034\\
		5&0.9693$\rpm$0.0044&0.9606$\rpm$0.0058&       \textbf{ 0.9587$\rpm$0.0053}$^{\uparrow\uparrow\uparrow}$&0.9429$\rpm$0.0056\\
		6&1$\rpm$0&0.99952$\rpm$0.00058                 &1$\rpm$0&1$\rpm$0\\
		7&0.784$\rpm$0.013&0.792$\rpm$0.015&            0.796$\rpm$0.014&0.776$\rpm$0.011\\
		8&0.9361$\rpm$0.0029&0.9346$\rpm$0.0037&        0.99691$\rpm$0.00084&0.99582$\rpm$0.00098\\
		9&0.8662$\rpm$0.0056&0.8659$\rpm$0.0053&       \textbf{ 0.8584$\rpm$0.004}$^{\uparrow\uparrow}$&0.8488$\rpm$0.0051\\
		10&0.6031$\rpm$0.0022&0.6$\rpm$0.0018&         \textbf{ 0.2873$\rpm$0.0056}$^{\uparrow\uparrow\uparrow}$&0.2114$\rpm$0.0013\\
		11&0.8855$\rpm$0.0041&\textbf{0.8944$\rpm$0.0037}$^{\uparrow}$&       0.8995$\rpm$0.0053&\textbf{0.9084$\rpm$0.0039}$^{\uparrow}$\\
		12&0.7454$\rpm$0.0064&0.747$\rpm$0.0085&        \textbf{0.711$\rpm$0.0063}$^{\uparrow}$&0.6949$\rpm$0.0076\\
		13&0.8614$\rpm$0.003&0.8628$\rpm$0.0036&        0.573$\rpm$0.006&\textbf{0.7359$\rpm$0.0034}$^{\uparrow\uparrow\uparrow}$\\
		14&0.2587$\rpm$0.0022&\textbf{0.2688$\rpm$0.0029}$^{\uparrow\uparrow\uparrow}$&       0.2317$\rpm$0.0028&\textbf{0.2661$\rpm$0.0029}$^{\uparrow\uparrow\uparrow}$\\
		15&0.372$\rpm$0.019&\textbf{0.496$\rpm$0.017}$^{\uparrow\uparrow\uparrow}$&           0.312$\rpm$0.015&\textbf{0.371$\rpm$0.015}$^{\uparrow\uparrow}$\\
		16&0.34$\rpm$0.018&\textbf{0.392$\rpm$0.015}$^{\uparrow}$&            0.1495$\rpm$0.0096&\textbf{0.269$\rpm$0.018}$^{\uparrow\uparrow\uparrow}$\\
		17&0.422$\rpm$0.012&\textbf{0.517$\rpm$0.017}$^{\uparrow\uparrow\uparrow}$&           0.394$\rpm$0.015&0.361$\rpm$0.014\\
		18&0.93778$\rpm$0.00073&0.93855$\rpm$0.00035&   0.7174$\rpm$0.004&0.7186$\rpm$0.0022\\
		19&0.8807$\rpm$0.0021&0.8784$\rpm$0.0018&       0.8514$\rpm$0.0033&0.8549$\rpm$0.0015\\
		\hline
	\end{tabular}
	
	\caption{Comparison of Hierarchical Multi-class Classification trained on a hierarchy obtained with the Best of 50 heuristic and on a hierarchy using Representative Based Dissimilarity. CART and logistic regression are used as learning algorithm in the nodes. Highlighted are the results that are significantly better than the rest. One arrow indicates a p-value of 0.05; two arrows, 0.01; three arrows, 0.001.}
	\label{table:Bo50_RBED_CART_logi}
\end{table}

In Table \ref{table:Bo50_RBED_CART_logi} using CART, we observe 8 datasets where there is no significant improvement over the random hierarchy (Table \ref{table:random_RBED_CART_logi}). In 5 out of the other 11 datasets, we don't observe significant differences between Bo50 and RBD approaches. We can see significant differences in favor of RBD in 5 datasets and in favor of Bo50 in 1 dataset. The most significant differences happen in datasets 14, 15, 16, 17, those that have 100 classes. Bo50 provides competitive results, especially if the number of classes is not very high. However, for those datasets with a high number of classes, it significantly underperforms when compared to RBD, and in some cases (15, 16, and 17), it does not provide results significantly better than the random hierarchy. As the number of classes increases, so does the number of possible hierarchies. It becomes less and less likely to sample a good hierarchy randomly.

In Table \ref{table:Bo50_RBED_CART_logi} using logistic regression, we observe just one dataset where there is no significant improvement over the random hierarchy (Table \ref{table:random_RBED_CART_logi}). We observe significant differences in favor of Bo50 in 9 datasets and in 5 datasets in favor of RBD. These 5 datasets are the same 5 datasets where we observed the same phenomenon using CART as the base classifier. In 8 of the 9 datasets where Bo50 significantly outperforms RBD using logistic regression, we did not observe any difference when using CART. Hierarchies that are useful for logistic regression are irrelevant for CART, arguably a more powerful classifier \citep{aslan2009}.

\subsection{Representative Based Dissimilarity vs Classifier Based Dissimilarity}
\label{sec:RBD vs CBD}

\begin{sidewaystable}%[t]
	\vspace{85mm}
	\centering
	\begin{tabular}{l|c|c|c|c|c|c|c|}
		\cline{2-8}
		&Random  & RBD  & CBD1 & CBD2 & CBD3 & OVA CART & OVA GLM\\
		\hline
		1&0.198$\rpm$0.019&0.248$\rpm$0.016&0.245$\rpm$0.015&0.245$\rpm$0.015&0.245$\rpm$0.015&\textbf{0.253$\rpm$0.019}&0.243$\rpm$0.013\\
        2&0.832$\rpm$0.015&0.9184$\rpm$0.0027&0.9108$\rpm$0.0035&0.894$\rpm$0.011&0.9332$\rpm$0.0052&0.9407$\rpm$0.002&\textbf{0.9446$\rpm$0.003}\\
        3&0.2264$\rpm$0.0096&0.2448$\rpm$0.007&0.243$\rpm$0.0078&\textbf{0.2714$\rpm$0.008}&0.2707$\rpm$0.0078&0.2681$\rpm$0.0073&0.2649$\rpm$0.0076\\
        4&0.486$\rpm$0.021&0.507$\rpm$0.034&0.536$\rpm$0.02&0.536$\rpm$0.02&0.536$\rpm$0.02&0.544$\rpm$0.024&\textbf{0.556$\rpm$0.026}\\
        5&0.86$\rpm$0.019&0.9429$\rpm$0.0056&\textbf{0.9587$\rpm$0.0052}&\textbf{0.9587$\rpm$0.0052}&\textbf{0.9587$\rpm$0.0052}&0.955$\rpm$0.0052&0.9574$\rpm$0.0052\\
        6&1$\rpm$0&1$\rpm$0&1$\rpm$0&1$\rpm$0&1$\rpm$0&1$\rpm$0&1$\rpm$0\\
        7&0.69$\rpm$0.021&0.776$\rpm$0.011&0.748$\rpm$0.011&\textbf{0.778$\rpm$0.011}&\textbf{0.778$\rpm$0.011}&0.773$\rpm$0.011&0.775$\rpm$0.012\\
        8&0.9619$\rpm$0.0072&0.99582$\rpm$0.00098&0.9705$\rpm$0.0068&0.99655$\rpm$8e-04&0.99627$\rpm$0.00075&0.99664$\rpm$0.00058&\textbf{0.99745$\rpm$0.00093}\\
        9&0.805$\rpm$0.019&0.8488$\rpm$0.0051&0.8337$\rpm$0.0045&0.8566$\rpm$0.0037&\textbf{0.8583$\rpm$0.0042}&0.8566$\rpm$0.004&0.8577$\rpm$0.0036\\
        10&0.2141$\rpm$0.0092&0.2114$\rpm$0.0013&0.20537$\rpm$9e-04&0.2081$\rpm$0.0053&0.2162$\rpm$0.0014&0.2129$\rpm$0.0013&\textbf{0.2326$\rpm$0.0013}\\
        11&0.832$\rpm$0.012&\textbf{0.9084$\rpm$0.0039}&0.8696$\rpm$0.0047&0.864$\rpm$0.011&0.849$\rpm$0.011&0.899$\rpm$0.01&0.9013$\rpm$0.0044\\
        12&0.627$\rpm$0.011&0.6949$\rpm$0.0076&0.6514$\rpm$0.0058&0.688$\rpm$0.011&0.685$\rpm$0.011&0.6993$\rpm$0.0071&\textbf{0.7189$\rpm$0.0072}\\
        13&0.451$\rpm$0.019&\textbf{0.7359$\rpm$0.0034}&0.404$\rpm$0.0056&0.6849$\rpm$0.006&0.7016$\rpm$0.0068&0.6764$\rpm$0.0078&0.6986$\rpm$0.005\\
        14&0.214$\rpm$0.0036&\textbf{0.2661$\rpm$0.0029}&0.1901$\rpm$0.0016&0.2298$\rpm$0.0038&0.2532$\rpm$0.0076&0.2476$\rpm$0.0071&0.2615$\rpm$0.0057\\
        15&0.204$\rpm$0.017&0.371$\rpm$0.015&0.181$\rpm$0.013&0.235$\rpm$0.028&0.235$\rpm$0.028&0.326$\rpm$0.018&\textbf{0.436$\rpm$0.018}\\
        16&0.105$\rpm$0.011&\textbf{0.269$\rpm$0.018}&0.095$\rpm$0.0088&0.095$\rpm$0.0088&0.095$\rpm$0.0088&0.198$\rpm$0.013&0.262$\rpm$0.015\\
        17&0.239$\rpm$0.016&0.361$\rpm$0.014&0.234$\rpm$0.017&0.234$\rpm$0.017&0.234$\rpm$0.017&0.366$\rpm$0.019&\textbf{0.536$\rpm$0.014}\\
        18&0.7022$\rpm$0.0045&\textbf{0.7186$\rpm$0.002}2&0.702$\rpm$0.0058&0.697$\rpm$0.036&0.7074$\rpm$0.0056&0.7074$\rpm$0.0056&0.7154$\rpm$0.0024\\
        19&0.767$\rpm$0.01&0.8549$\rpm$0.0015&0.8161$\rpm$0.0012&0.8524$\rpm$0.0039&0.8564$\rpm$0.0022&0.8597$\rpm$0.0018&\textbf{0.8605$\rpm$0.0016}\\
		\hline
	\end{tabular}
	\caption{Comparison of Hierarchical Multi-class Classification trained on different hierarchies: random hierarchy, RBD hierarchy, Flat Classifier CBD (increasing complexity), and One-vs-all CBD (using CART and logistic regression (GLM) as learning algorithms). Highlighted are the highest accuracy for each dataset, which does not necessarily mean that it is significantly better.}
	\label{table:rbed_cbd}
\end{sidewaystable}

Another popular approach to measure dissimilarities between classes is using an initial classifier to evaluate how similar classes are to each other. \textit{Classifier Based Dissimilarity} builds on the concept of separability and measures how well a classifier can separate between the different classes. In these experiments, we are going to compare CBD and RBD. While RBD is a deterministic metric, CBD will depend on how we train the classifier. We will measure how the quality of the CBD classifier affects the quality of the hierarchy, and therefore, the classification performance. 

To obtain the Classifier Based Dissimilarity, we are going to run single classifiers (using CART with different complexities) and OVA (using CART and GLM as binary classifiers). For the single classifiers we use CART with three different sets of complexity parameters: $(0.5,0.1),$ $(0.5,0.1,0.05,0.01),$ $(0.5,0.1,0.05,0.01,0.005,0.001)$. For OVA using CART we select the complexity parameter from the values $(0.5,0.1,0.05,0.01,0.005,0.001)$ using 10 fold cross-validation for each of the binary classifiers. Once we have the CBD classifiers, we compute the dissimilarity as explained in \ref{subsec:avatesting}. To magnify the effects of the hierarchy's quality, we are going to use logistic regression as a learning algorithm on the hierarchy. 

\begin{figure}
	\centering
	\begin{minipage}{.5\textwidth}
		\centering
		\includegraphics[width=1\linewidth]{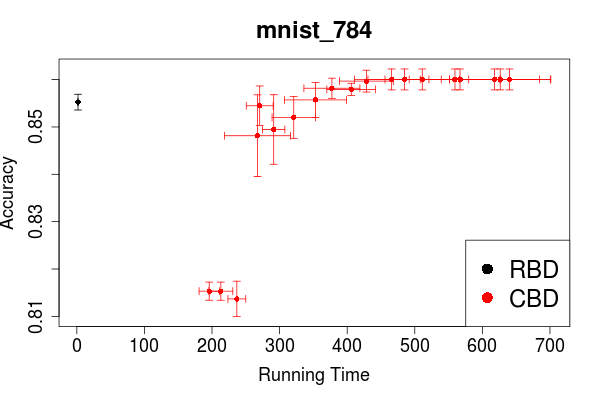}
	\end{minipage}%
	\begin{minipage}{.5\textwidth}
		\centering
		\includegraphics[width=1\linewidth]{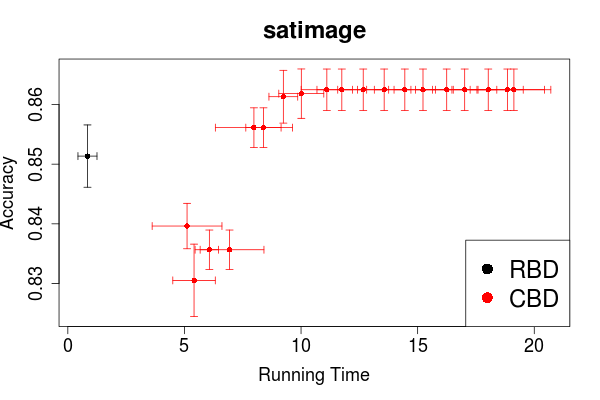}
	\end{minipage}
	\caption{Evolution of the performance of Hierarchical Multi-class Classification trained on different hierarchies as the complexity of the CBD classifier increases, compared against RBD.}
	\label{fig:rtime_acc}
\end{figure}

Results are presented in Table \ref{table:rbed_cbd}. In 9 out of 19 datasets, at least one CBD variant presents significantly better results than RBD. The counterexample only happens in one dataset.

In some cases the hierarchies using the simplest CBD classifier outperforms its RBD counterpart; in others, hierarchies with simple CBD classifiers perform worse than RBD, but this difference disappears as the CBD classifier becomes more complex. OVA using logistic regression as CBD classifier produces significantly better results than OVA using CART as CBD classifier in 6 datasets, while the opposite happens in only one dataset. This is especially true in those datasets with more classes. The learning algorithm used on the hierarchy is also logistic regression; this opens interesting questions about the relationship between the CBD classifier and the base classifier trained on the hierarchy.

In general, we observe an improvement as the complexity of the CBD classifier increases. In Figure \ref{fig:rtime_acc}, we can see a finer detail on this effect. Similar to the CBD columns in Table \ref{table:rbed_cbd}, we have chosen different configurations of the CART algorithm's complexity parameter, increasing the allowed complexity of the CBD classifiers. As a proxy for the complexity, we have calculated the running time it takes to train the CBD classifier. Clearly, the range of classifiers complexities explored in this experiment is very limited. Intuitively, however, it is obvious that CBD cannot be thought of as a single measure and, as such, compared directly against RBD. For any given dataset, it is possible to find a classifier that will create a worse, equal, or better hierarchy than RBD (if RBD has not found the best possible hierarchy). CBD is a family of dissimilarities and needs to be treated as such.

In \citep{Silva-palacios2018}, a direct comparison is performed between a version of CBD and RBD without acknowledging all the different possibilities to obtain CBD dissimilarities. In \citep{Godbole2002}, \citep{Chennupati2016}, single versions of CBD are used to extract hierarchies. The results are compared with non-hierarchical approaches. In \citep{Bengio2010} and \citep{mnih2009}, single versions of CBD are introduced to extract hierarchies. The results are compared with random hierarchies. To the best of our knowledge, there has not been a deep study of the influence on how CBD is obtained and the final classification performance of the hierarchical classification problem.

OVA using logistic regression as CBD classifier produces significantly better results than OVA using CART as CBD classifier in 6 datasets, while the opposite happens in only one dataset. This is especially true in those datasets with more classes. The learning algorithm used on the hierarchy is also logistic regression; this opens interesting questions about the relationship between the CBD classifier and the base classifier trained on the hierarchy.

\subsection{Relationship between the base classifier and the CBD classifier}

\begin{figure}[!ht]
	\centering
	\subfloat[][]{\includegraphics[width=.4\textwidth]{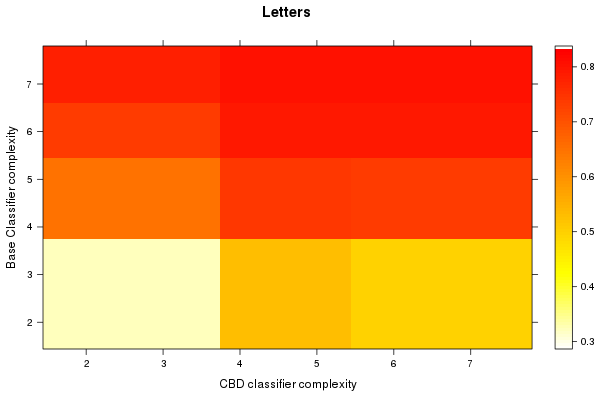}}\quad
	\subfloat[][]{\includegraphics[width=.4\textwidth]{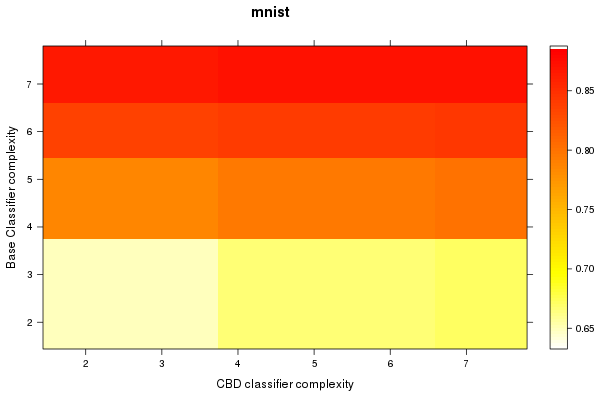}}\\
	\subfloat[][]{\includegraphics[width=.4\textwidth]{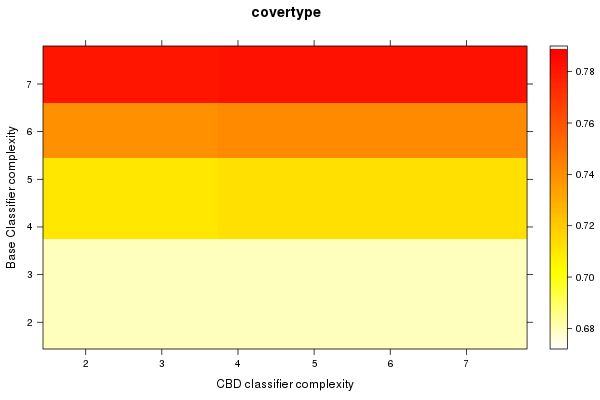}}\quad
	\subfloat[][]{\includegraphics[width=.4\textwidth]{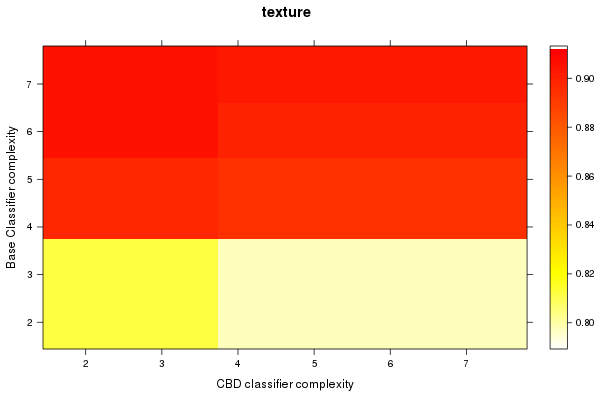}}
	\caption{Comparison of the complexity of the CBD classifier and the base classifier}
	\label{fig:heatmap}
\end{figure}

In the previous experiment, we have established that CBD using OVA and logistic regression performs better than CBD using CART when we also use logistic regression as base classifier. The intention behind CBD is to measure how easy two classes are to separate. More specifically, we want to know how easy it will be for the learning algorithm trained on the hierarchy to separate between these two classes. However, the results of training the CBD classifier give us only an approximation of how well the base classifier will perform. 

With this motivation in mind, in this experiment, we want to explore the relationship between the power of the CBD classifier, used to measure dissimilarities between classes, and the power of the base classifier trained on the hierarchy. The same experimental setup from the previous experiments applies. As CBD classifier, we use a single CART as in the previous section. As the base classifier, we also choose CART.

Figure \ref{fig:heatmap} shows how the classification accuracy across four different datasets changes depending on choosing different configurations of the complexity parameters for both the CBD classifier and the base classifier. The most interesting results can be seen in Figure \ref{fig:heatmap}a: for the simpler base classifier, the accuracy drops as the CBD classifier gets more complex. The same behaviour can be seen in \ref{fig:heatmap}d, but not in \ref{fig:heatmap}b and \ref{fig:heatmap}c. This shows that the relationships between classes found by a powerful CBD classifier will not necessarily be useful when we have a weak base classifier. 

In all four examples in Figure \ref{fig:heatmap}, the relevance of the quality of the hierarchy decreases as the quality of the base classifier increases. This confirms our conclusion that a powerful enough base classifier is insensitive to the quality of the hierarchy. 

Finally, in Figure \ref{fig:heatmap}c, one can see that there are very small differences in overall classification performance as the quality of the hierarchy increases for any configuration. This suggests that for this dataset, there is no inherent structure between classes.

\subsection{Hierarchical Agglomerative Clustering vs. Hierarchical K-Means}
\label{sec:HAC_HKM}

\begin{table}[t]
	\centering
	\begin{tabular}{l|c|c||c|c|}
		% \cline{2-7}
		& \multicolumn{4}{|c|}{CART} \\
		\cline{2-5}
		&RBD-HAC&RBD-HKM&OVA-HAC&OVA-HKM\\
		\hline
		1&0.171$\rpm$0.016&0.173$\rpm$0.014&0.151$\rpm$0.018&0.16$\rpm$0.017\\
		2&0.9558$\rpm$0.0024&0.9506$\rpm$0.0019&0.9585$\rpm$0.0026&0.9588$\rpm$0.0032\\
		3&0.229$\rpm$0.011&0.236$\rpm$0.011&0.2467$\rpm$0.0092&0.257$\rpm$0.01\\
		4&0.767$\rpm$0.019&0.749$\rpm$0.015&0.72$\rpm$0.025&0.684$\rpm$0.029\\
		5&0.9606$\rpm$0.0058&0.9621$\rpm$0.0043&0.9673$\rpm$0.0045&0.9658$\rpm$0.0042\\
		6&0.99952$\rpm$0.00058&0.9971$\rpm$0.0017&1$\rpm$0&1$\rpm$0\\
		7&0.792$\rpm$0.015&0.786$\rpm$0.018&0.781$\rpm$0.013&0.769$\rpm$0.015\\
		8&0.9346$\rpm$0.0037&0.9365$\rpm$0.0035&0.935$\rpm$0.0026&0.9309$\rpm$0.0028\\
		9&0.8659$\rpm$0.0053&0.8631$\rpm$0.0056&0.8613$\rpm$0.0048&0.8662$\rpm$0.0054\\
		10&0.6$\rpm$0.0018&0.6047$\rpm$0.0019&0.605$\rpm$0.0013&0.6023$\rpm$0.0015\\
		11&0.8944$\rpm$0.0037&0.8838$\rpm$0.0041&0.8834$\rpm$0.0074&0.8876$\rpm$0.0048\\
		12&0.747$\rpm$0.0085&0.7435$\rpm$0.0061&0.73$\rpm$0.0071&0.7397$\rpm$0.0069\\
		13&0.8628$\rpm$0.0036&0.8666$\rpm$0.003&0.8673$\rpm$0.0037&0.8651$\rpm$0.0029\\
		14&0.2688$\rpm$0.0029&0.2704$\rpm$0.0019&0.2558$\rpm$0.0042&\textbf{0.2754$\rpm$0.0023}$^{\uparrow\uparrow\uparrow}$\\
		15&0.496$\rpm$0.017&0.466$\rpm$0.018&0.42$\rpm$0.017&0.381$\rpm$0.02\\
		16&0.392$\rpm$0.015&0.411$\rpm$0.018&0.367$\rpm$0.023&0.365$\rpm$0.017\\
		17&0.517$\rpm$0.017&0.484$\rpm$0.016&\textbf{0.474$\rpm$0.015}$^{\uparrow\uparrow}$&0.418$\rpm$0.017\\
		18&0.93855$\rpm$0.00035&0.93854$\rpm$0.00035&0.93851$\rpm$0.00037&0.93852$\rpm$3e-04\\
		19&0.8784$\rpm$0.0018&\textbf{0.8856$\rpm$0.0017}$^{\uparrow\uparrow\uparrow}$&\textbf{0.8857$\rpm$0.0017}$^{\uparrow\uparrow}$&0.8808$\rpm$0.002\\
		\hline
		depths & 9.2 & 6.6 & 13.5 & 6.5\\
		\hline
	\end{tabular}
	
	\caption{Comparison HAC and HKM using CART as learning algorithm}
	\label{table:HACHKMr}
\end{table}

So far, we have only discussed differences in the hierarchies depending on how we measure dissimilarity between classes. In this experiment, we explore the differences created by different clustering algorithms. In particular, we will focus on two hierarchical clustering algorithms: Hierarchical Agglomerative Clustering (HAC) and Hierarchical K-Means (HKM). 

The experimental setup is similar to the previous sections. In Tables \ref{table:HACHKMr} and \ref{table:HACHKMg}, we can see the results of using CART and logistic regression as learning algorithms. In both cases, we will compare the clustering algorithms using as dissimilarity metric RBD and CBD using OVA. When we use CART as a base learner, we will also use CART as a classifier for OVA; similarly, we will use logistic regression for OVA when using logistic regression as a base classifier.

Using CART as a base classifier, we cannot observe many significant differences between HAC and HKM in the pairwise comparisons. The main improvement in the hierarchies comes from the information shown in the dissimilarities between classes and not by the clustering algorithm.

\begin{figure}
	\centering
	\begin{minipage}{.5\textwidth}
		\centering
		\includegraphics[width=1\linewidth]{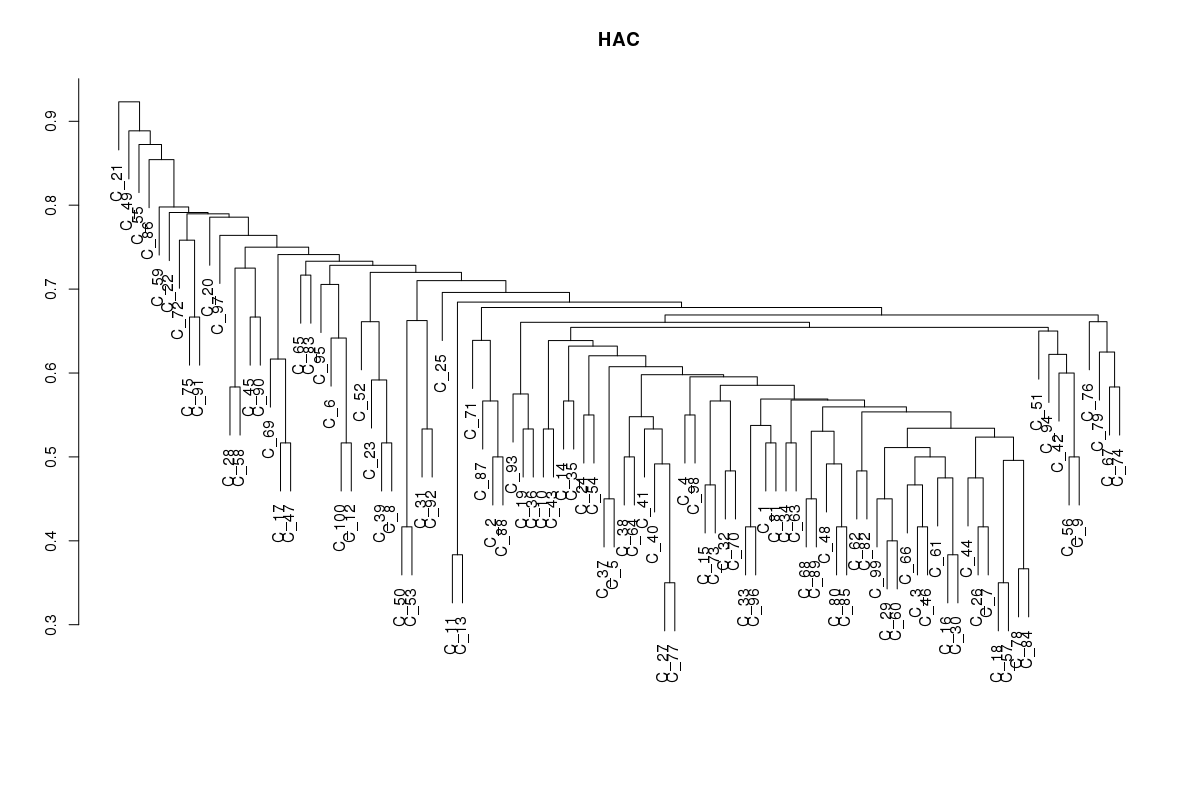}
	\end{minipage}%
	\begin{minipage}{.5\textwidth}
		\centering
		\includegraphics[width=1\linewidth]{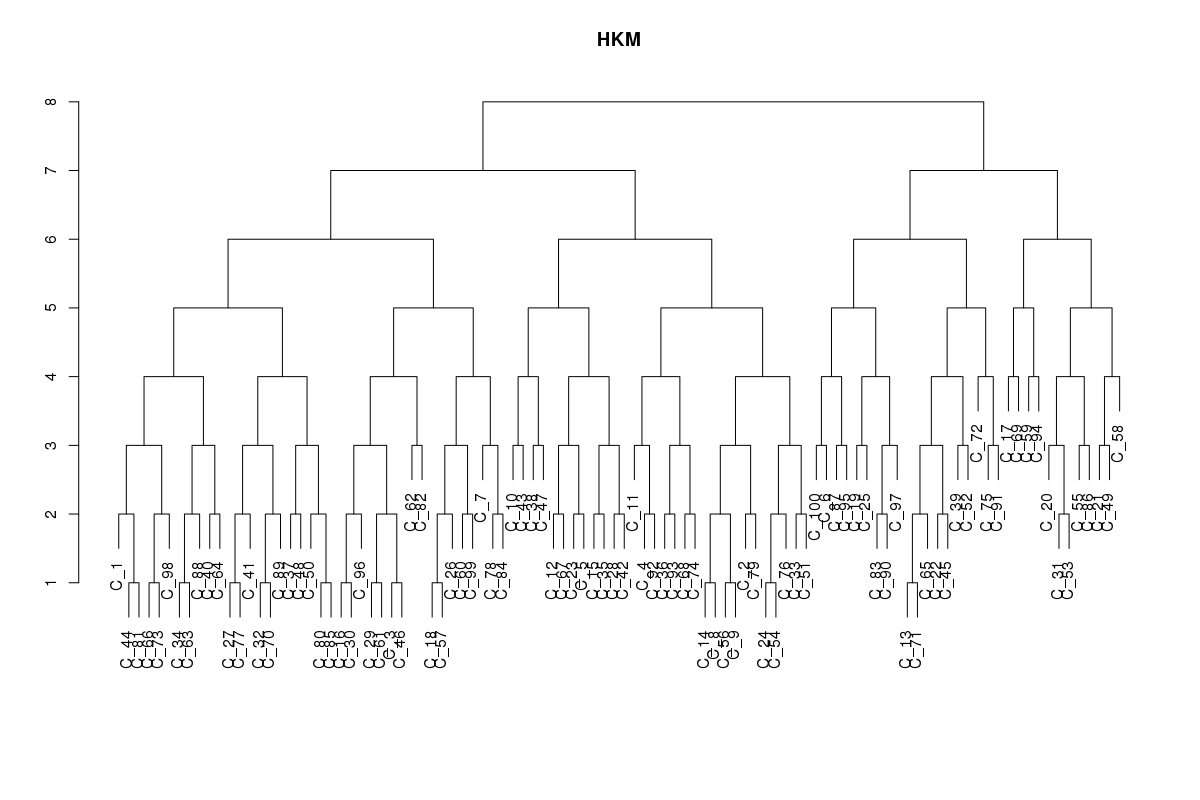}
	\end{minipage}
	\caption{On the left, hierarchy obtained with Hierarchical Agglomerative Clustering with average link. On the right, hierarchy obtained with Hierarchical K-means.}
	\label{fig:h_t_c}
\end{figure}

On the other hand, if we use logistic regression as a base classifier, we find several significant differences in the pairwise comparisons in favor of HMC. One reason to explain this is the shape of the final hierarchy and the choice of logistic regression as a learner. As suggested by the average depths of the hierarchies created, HAC tends to create hierarchies in a chain-shape, while HKM tends to generate more balanced hierarchies (see Figure \ref{fig:h_t_c} for an example).

Depending on the learning algorithm chosen, the final results might vary not only based on the information about how similar the classes are, but also on the particular shape of the hierarchy. It is important to take this into account when designing future experiments.

\begin{table}[t]
	\centering
	\begin{tabular}{l|c|c||c|c|}
		% \cline{2-7}
		& \multicolumn{4}{|c|}{Logistic Regression} \\
		\cline{2-5}
		&RBD-HAC&RBD-HKM&OVA-HAC&OVA-HKM\\
		\hline
		1&0.248$\rpm$0.016&0.25$\rpm$0.016&0.243$\rpm$0.013&0.265$\rpm$0.016\\
		2&\textbf{0.9184$\rpm$0.0027}$^{\uparrow\uparrow\uparrow}$&0.887$\rpm$0.0042&\textbf{0.9446$\rpm$0.003}$^{\uparrow\uparrow\uparrow}$&0.9167$\rpm$0.0034\\
		3&0.2448$\rpm$0.007&0.243$\rpm$0.01&0.2649$\rpm$0.0076&0.2712$\rpm$0.0079\\
		4&0.507$\rpm$0.034&0.458$\rpm$0.03&0.556$\rpm$0.026&0.536$\rpm$0.037\\
		5&0.9429$\rpm$0.0056&0.9429$\rpm$0.0061&\textbf{0.9574$\rpm$0.0052}$^{\uparrow\uparrow\uparrow}$&0.8645$\rpm$0.0074\\
		6&1$\rpm$0&1$\rpm$0&1$\rpm$0&1$\rpm$0\\
		7&0.776$\rpm$0.011&0.786$\rpm$0.013&0.775$\rpm$0.012&0.791$\rpm$0.011\\
		8&0.99582$\rpm$0.00098&0.99555$\rpm$0.00078&\textbf{0.99745$\rpm$0.00093}$^{\uparrow\uparrow\uparrow}$&0.9884$\rpm$0.0021\\
		9&0.8488$\rpm$0.0051&0.8488$\rpm$0.0051&0.8577$\rpm$0.0036&0.8593$\rpm$0.0036\\
		10&0.2114$\rpm$0.0013&\textbf{0.2194$\rpm$0.0013}$^{\uparrow\uparrow\uparrow}$&0.2326$\rpm$0.0013&\textbf{0.2444$\rpm$0.002}$^{\uparrow\uparrow\uparrow}$\\
		11&\textbf{0.9084$\rpm$0.0039}$^{\uparrow\uparrow\uparrow}$&0.8887$\rpm$0.0041&\textbf{0.9013$\rpm$0.0044}$^{\uparrow\uparrow\uparrow}$&0.8693$\rpm$0.0035\\
		12&0.6949$\rpm$0.0076&0.7108$\rpm$0.0095&0.7189$\rpm$0.0072&0.7111$\rpm$0.0066\\
		13&\textbf{0.7359$\rpm$0.0034}$^{\uparrow\uparrow\uparrow}$&0.6916$\rpm$0.0045&\textbf{0.6986$\rpm$0.005}$^{\uparrow\uparrow\uparrow}$&0.6638$\rpm$0.0058\\
		14&0.2661$\rpm$0.0029&\textbf{0.2767$\rpm$0.0024}$^{\uparrow\uparrow\uparrow}$&0.2615$\rpm$0.0057&0.2628$\rpm$0.0023\\
		15&\textbf{0.371$\rpm$0.015}$^{\uparrow\uparrow}$&0.327$\rpm$0.014&\textbf{0.436$\rpm$0.018}$^{\uparrow\uparrow\uparrow}$&0.236$\rpm$0.019\\
		16&\textbf{0.269$\rpm$0.018}$^{\uparrow\uparrow}$&0.228$\rpm$0.017&\textbf{0.262$\rpm$0.015}$^{\uparrow\uparrow\uparrow}$&0.14$\rpm$0.012\\
		17&\textbf{0.361$\rpm$0.014}$^{\uparrow\uparrow}$&0.298$\rpm$0.016&\textbf{0.536$\rpm$0.014}$^{\uparrow\uparrow\uparrow}$&0.267$\rpm$0.011\\
		18&0.7186$\rpm$0.0022&0.697$\rpm$0.021&0.7154$\rpm$0.0024&0.698$\rpm$0.02\\
		19&0.8549$\rpm$0.0015&\textbf{0.8661$\rpm$0.0019}$^{\uparrow\uparrow\uparrow}$&\textbf{0.8605$\rpm$0.0016}$^{\uparrow\uparrow\uparrow}$&0.8457$\rpm$0.0033\\
		\hline
		depths & 9.2 & 6.6 & 18.1 & 6.7\\
		\hline
	\end{tabular}
	
	\caption{Comparison HAC and HKM using logistic regression as learning algorithm}
	\label{table:HACHKMg}
\end{table}

\section{Conclusions and Discussions.}
\label{sec:Conclusion}

Multi-class classification is the task of classifying instances into more than two classes. The most popular approaches to solve multi-class problems (All-vs-all, One-vs-all, extended algorithms) become prohibitively expensive and time-consuming as the number of classes increases. Hierarchical Multi-class Classification (HMC) significantly reduces the time needed for predicting the class of a new instance while maintaining reasonable training time and classification performance \citep{Bengio2010}.

HMC significantly reduces testing time by creating a hierarchy of classes and reducing the number of classifiers that need to be evaluated to produce a new prediction. There are several solutions proposed in the literature, but this paper is the first comprehensive discussion on what is a good hierarchy and how to measure its relevance. We demonstrate that the relevance of the hierarchy's quality depends on the trade-off between the complexity of the classification problem and the complexity of the learning algorithm used. We have compared the state of the art practices to extract hierarchies and evaluated them. In the process, we have identified some of the common pitfalls of extracting hierarchies and how to avoid them for HMC.

We have confirmed that the quality of the hierarchy does not always affect the performance of the classifier. A hierarchy will only be useful if it can exploit the existing relationships between classes. This requires that there is an actual structure in how classes relate to each other and that the extraction method can find it -- and neither of those assumptions necessarily holds for all datasets.

In addition, it requires that the extracted hierarchy is useful for the learning algorithm. A useful hierarchy isolates easier-to-find boundaries between classes at the nodes closer to the dendrogram's root, while harder-to-find boundaries are left for the nodes near the leaves. Whether the structure in the classes creates easier or harder to find decision boundaries will depend on the learning algorithm chosen as the base classifier.

We have shown that the weaker the base classifier is, the more relevant the hierarchy's quality becomes. On the other hand, we have observed that for some datasets, a hierarchy can be useful using a simple classifier as the base classifier but irrelevant for a more powerful classifier. For any given dataset, a powerful enough classifier will not be affected by the hierarchy's quality. The trade-off between the complexity of the classification problem and the classifiers' predictive power needs to be taken into account to evaluate the goodness of a method to extract hierarchies. 

All of this means that in many common settings, the hierarchy's quality becomes irrelevant for the overall predictive performance. At the same time, the divide-and-conquer aspect inherent in the HMC approach may provide benefits over alternatives, like either single classifiers or OVA. Ignoring this might lead practitioners to make claims about the quality of a hierarchy when, in fact, they are validating other aspects. To avoid these cases, we suggest using HMC on a random hierarchy as an appropriate benchmark to understand the extracted hierarchy's quality. By comparing with a random hierarchy, we can isolate the added value of the hierarchy's quality. If there is no existing relationship between classes relevant to the learning algorithm chosen, extracted hierarchies will not provide results significantly better than those coming from random hierarchies. 

\begin{figure}
	\centering
	\begin{minipage}{.5\textwidth}
		\centering
		\includegraphics[width=1\linewidth]{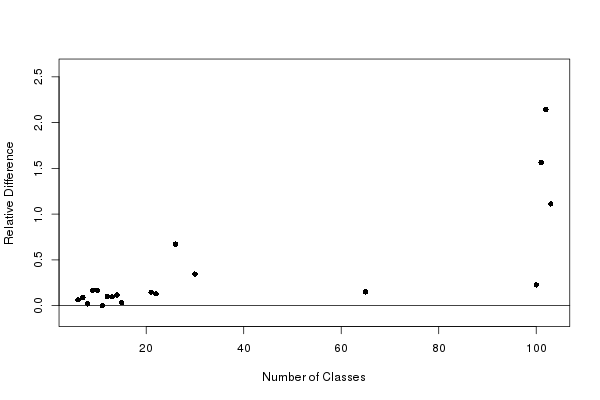}
	\end{minipage}%
	\begin{minipage}{.5\textwidth}
		\centering
		\includegraphics[width=1\linewidth]{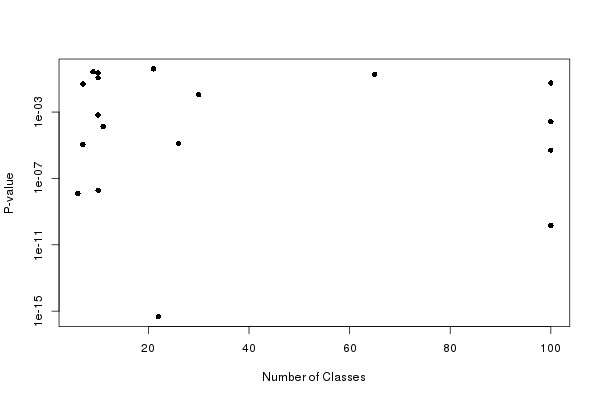}
	\end{minipage}
	\caption{On the left, relative difference of accuracies as the number of classes increases, measured with respect to the random hierarchy. On the right, significance of the difference between the random hierarchy and the best informed hierarchy.}
	\label{fig:classes}
\end{figure}

All along with our experiments, we have observed that the hierarchy's quality becomes more and more relevant as the number of classes increases, as shown in Figure \ref{fig:classes}. This result is intuitive: as the number of classes increases, it is natural to expect that more complex structures exist between classes and good hierarchies become crucial, at least as long as we are using a fixed base classifier. 

Figure \ref{fig:classes} shows the resulting p-values of comparing the results of the best hierarchy extracted and the random one. Some of the values are so extreme that they lose their significance since our experiments are only considering 20 repetitions of sampling training and test sets. However, it shows that the structures found by our informed hierarchies do exist and that the chances of obtaining them randomly can be quite low.

We have found hints that suggest that some base classifiers are more sensitive to the shape of the hierarchy (whether it has a tree-shape or a chain-shape). Hierarchical clustering with average link tends to provide deeper dendrograms than hierarchical K-means. In our case, using logistic regression as the base classifier, we have found significant differences in favor of the chain-shape structures. However, using CART as the base classifier, we have not found apparent differences between HAC and K-means.

Comparing different clustering algorithms just based on the classification performance gives only a partial view of the problem. Without considering how the learning algorithm used is affected by the shape of the dendrogram might lead practitioners to misleading conclusions. The relationship between these shapes and the requirements of testing times remains for future work.

We have found that the hierarchy's quality depends decisively on how we measure the dissimilarity between classes. We have reviewed two existing approaches: representative-based dissimilarities (RBD) and classifier-based dissimilarity (RBD). Using RBD is fast and simple and often produces significant improvements when compared to random hierarchies. However, we have found that hierarchies obtained with CBD dissimilarities are in general as good or better. 

We have also evaluated the hierarchies obtained using the Best of 50 heuristic presented in \citep{melnikov2018}. This method samples 50 random hierarchies and picks the best one using a cross-validation scheme. When the number of classes is not too high, this method presents very competitive results, outperforming in some cases RBD and CBD hierarchies. There are some relationships between the classes that these dissimilarities or the clustering algorithm, are failing to measure and encode in a hierarchy.

However, for datasets with many classes, Best of 50 underperforms compared to CBD or RBD hierarchies. The possible number of hierarchies increases very fast with the number of classes. Best of 50 relies on the assumption that the sampled hierarchies are representative of the general distribution of all possible hierarchies; this assumption can break as the number of classes increases.

CBD is a family of dissimilarities that depend on the classifiers used. It is usually accepted in the literature that a simple classifier used to measure CBD is good enough to obtain a good hierarchy. We have proven that this is not necessarily the case. While we can get better hierarchies with the appropriate CBD dissimilarity, RBD can outperform too simple CBD hierarchies. 

We have analyzed the trade-off between the complexity of the CBD classifier and the base classifiers trained on the hierarchy's nodes. Our results show that the hierarchy's quality is more relevant for the weaker base classifiers than for the more powerful ones. An interesting result is how hierarchies using simpler CBD classifiers outperform the ones using complex CBD classifiers when the base classifier is weak. This opens up interesting research lines into understanding the synergies between the quality of the extracted hierarchies and the power of the base classifiers used for HMC.

Throughout this paper, we have evaluated the quality of the hierarchies based on the results of a classifier trained on them. This evaluation depends on the particular characteristics of the datasets used, but also on the learning algorithms used on the hierarchies. With a powerful enough classifier, the hierarchy's quality becomes irrelevant, i.e., all hierarchies are equally good. This suggests that there must be more appropriate ways of measuring the hierarchy's quality, in case one is explicitly interested in that. For example, the task of finding interesting relationships between classes in an unsupervised manner can be approached through finding a good hierarchy of classes, but HMC accuracy would not be a sufficient evaluation metric. Multi-class classification is the task of classifying instances into more than two classes. The most popular approaches to solve multi-class problems (All-vs-all, One-vs-all, extended algorithms) become prohibitively expensive and time-consuming as the number of classes increases. Hierarchical Multi-class Classification (HMC) significantly reduces the time needed for predicting the class of a new instance while maintaining reasonable training time and classification performance \citep{Bengio2010}.

HMC significantly reduces testing time by creating a hierarchy of classes and reducing the number of classifiers that need to be evaluated to produce a new prediction. There are several solutions proposed in the literature, but this paper is the first comprehensive discussion on what is a good hierarchy and how to measure its relevance. We demonstrate that the relevance of the hierarchy's quality depends on the trade-off between the complexity of the classification problem and the complexity of the learning algorithm used. We have compared the state of the art practices to extract hierarchies and evaluated them. In the process, we have identified some of the common pitfalls of extracting hierarchies and how to avoid them for HMC.

We have confirmed that the quality of the hierarchy does not always affect the performance of the classifier. A hierarchy will only be useful if it can exploit the existing relationships between classes. This requires an actual structure in how classes relate to each other and that the extraction method can find it -- and neither of those assumptions necessarily holds for all datasets.

In addition, it requires that the extracted hierarchy is useful for the learning algorithm. A useful hierarchy isolates easier-to-find boundaries between classes at the nodes closer to the dendrogram's root, while harder-to-find boundaries are left for the nodes near the leaves. Whether the structure in the classes creates easier or harder to find decision boundaries will depend on the learning algorithm chosen as the base classifier.

We have shown that the weaker the base classifier is, the more relevant the hierarchy's quality becomes. On the other hand, we have observed that for some datasets, a hierarchy can be useful using a simple classifier as the base classifier but irrelevant for a more powerful classifier. For any given dataset, a powerful enough classifier will not be affected by the hierarchy's quality. The trade-off between the complexity of the classification problem and the classifiers' predictive power needs to be taken into account to evaluate the goodness of a method to extract hierarchies.

\section*{Acknowledgments}

This work was supported by "Stiftelsen för kunskaps- och kompetensutveckling".


\begin{thebibliography}{10}
\expandafter\ifx\csname url\endcsname\relax
  \def\url#1{\texttt{#1}}\fi
\expandafter\ifx\csname urlprefix\endcsname\relax\def\urlprefix{URL }\fi
\expandafter\ifx\csname href\endcsname\relax
  \def\href#1#2{#2} \def\path#1{#1}\fi

\bibitem{aly2005}
M.~Aly, Survey on multiclass classification methods, Neural Netw 19 (2005)
  1--9.

\bibitem{Bengio2010}
S.~Bengio, J.~Weston, D.~Grangier, {Label embedding trees for large multi-class
  tasks}, Advances in Neural Information Processing Systems 23~(1) (2010)
  163--171.

\bibitem{Rifkin_2004}
R.~Rifkin, A.~Klautau, In defense of one-vs-all classification, J. Mach. Learn.
  Res. 5 (2004) 101--141.

\bibitem{stanley1986}
R.~P. Stanley, What is enumerative combinatorics?, in: Enumerative
  combinatorics, Springer, 1986, pp. 1--63.

\bibitem{Vural2004}
V.~Vural, J.~G. Dy, A hierarchical method for multi-class support vector
  machines, in: Proceedings of the Twenty-first International Conference on
  Machine Learning, ICML '04, ACM, New York, NY, USA, 2004, pp. 105--.

\bibitem{Silva-palacios2018}
D.~Silva-palacios, C.~Ferri, M.~J. Ram{\'{i}}rez-quintana, {Probabilistic class
  hierarchies for multiclass classification}, Journal of Computational Science
  (2018) 1--10.

\bibitem{Li2007}
T.~Li, S.~Zhu, M.~Ogihara, {Hierarchical document classification using
  automatically generated hierarchy}, Journal of Intelligent Information
  Systems 29~(2) (2007) 211--230.

\bibitem{Godbole2002}
S.~Godbole, {Exploiting confusion matrices for automatic generation of topic
  hierarchies and scaling up multi-way classifiers}, Progress Report, IIT
  Bombay (2002) 17.

\bibitem{weston1998}
J.~Weston, C.~Watkins, Multi-class support vector machines, Tech. rep.,
  Citeseer (1998).

\bibitem{lee2004}
Y.~Lee, Y.~Lin, G.~Wahba, Multicategory support vector machines: Theory and
  application to the classification of microarray data and satellite radiance
  data, Journal of the American Statistical Association 99~(465) (2004) 67--81.

\bibitem{engel1988}
J.~Engel, Polytomous logistic regression, Statistica Neerlandica 42~(4) (1988)
  233--252.

\bibitem{hastie1998}
T.~Hastie, R.~Tibshirani, Classification by pairwise coupling, in: Advances in
  neural information processing systems, 1998, pp. 507--513.

\bibitem{galar2011}
M.~Galar, A.~Fern{\'a}ndez, E.~Barrenechea, H.~Bustince, F.~Herrera, An
  overview of ensemble methods for binary classifiers in multi-class problems:
  Experimental study on one-vs-one and one-vs-all schemes, Pattern Recognition
  44~(8) (2011) 1761--1776.

\bibitem{liu2017}
W.~Liu, I.~W. Tsang, K.-R. M{\"u}ller, An easy-to-hard learning paradigm for
  multiple classes and multiple labels, The Journal of Machine Learning
  Research 18~(1) (2017) 3300--3337.

\bibitem{Babbar2013}
R.~Babbar, I.~Partalas, E.~Gaussier, M.~R. Amini, On flat versus hierarchical
  classification in large-scale taxonomies, in: Advances in Neural Information
  Processing Systems 26, Curran Associates, Inc., 2013, pp. 1824--1832.

\bibitem{Qu2017}
Y.~Qu, L.~Lin, F.~Shen, C.~Lu, Y.~Wu, Y.~Xie, D.~Tao, Joint hierarchical
  category structure learning and large-scale image classification, IEEE
  Transactions on Image Processing 26~(9) (2017) 4331--4346.

\bibitem{silla2011}
C.~N. Silla, A.~A. Freitas, A survey of hierarchical classification across
  different application domains, Data Mining and Knowledge Discovery 22~(1-2)
  (2011) 31--72.

\bibitem{Chen2004}
Y.~Chen, M.~M. Crawford, J.~Ghosh, Integrating support vector machines in a
  hierarchical output space decomposition framework, in: IGARSS 2004. 2004 IEEE
  International Geoscience and Remote Sensing Symposium, Vol.~2, 2004, pp.
  949--952 vol.2.

\bibitem{babbar2016}
R.~Babbar, I.~Partalas, E.~Gaussier, M.-R. Amini, C.~Amblard, Learning taxonomy
  adaptation in large-scale classification, The Journal of Machine Learning
  Research 17~(1) (2016) 3350--3386.

\bibitem{naik2019}
A.~Naik, H.~Rangwala, Improving large-scale hierarchical classification by
  rewiring: a data-driven filter based approach, Journal of Intelligent
  Information Systems 52~(1) (2019) 141--164.

\bibitem{morin2005}
F.~Morin, Y.~Bengio, Hierarchical probabilistic neural network language model.,
  in: Aistats, Vol.~5, Citeseer, 2005, pp. 246--252.

\bibitem{mnih2009}
A.~Mnih, G.~E. Hinton, A scalable hierarchical distributed language model, in:
  Advances in neural information processing systems, 2009, pp. 1081--1088.

\bibitem{Frank_2004}
E.~Frank, S.~Kramer, Ensembles of nested dichotomies for multi-class problems,
  in: Proceedings of the Twenty-first International Conference on Machine
  Learning, ICML '04, ACM, New York, NY, USA, 2004, pp. 39--.

\bibitem{melnikov2018}
V.~Melnikov, E.~H{\"u}llermeier, On the effectiveness of heuristics for
  learning nested dichotomies: an empirical analysis, Machine Learning
  107~(8-10) (2018) 1537--1560.

\bibitem{Larsen_1999}
B.~Larsen, C.~Aone, Fast and effective text mining using linear-time document
  clustering, in: Proceedings of the Fifth ACM SIGKDD International Conference
  on Knowledge Discovery and Data Mining, KDD '99, ACM, New York, NY, USA,
  1999, pp. 16--22.

\bibitem{Steinbach00}
M.~Steinbach, G.~Karypis, V.~Kumar, A comparison of document clustering
  techniques, in: In KDD Workshop on Text Mining, 2000.

\bibitem{Chennupati2016}
S.~Chennupati, S.~Sah, S.~Nooka, R.~Ptucha, {Hierarchical Decomposition of
  Large Deep Networks}, Electronic Imaging 2016~(19) (2016) 1--6.

\bibitem{Tsoumakas2008}
G.~Tsoumakas, I.~Katakis, I.~Vlahavas, {Effective and efficient multilabel
  classification in domains with large number of labels}, Proc. ECML/PKDD 2008
  Workshop on Mining Multidimensional Data (MMD'08) (2008) 30--44.

\bibitem{Nadeau2003}
C.~Nadeau, Y.~Bengio, {Inference for the generalization error}, Machine
  Learning 52~(3) (2003) 239--281.

\bibitem{aslan2009}
O.~{Asian}, O.~T. {Yildiz}, E.~{Alpaydin}, Calculating the vc-dimension of
  decision trees, in: 2009 24th International Symposium on Computer and
  Information Sciences, 2009, pp. 193--198.
\newblock \href {https://doi.org/10.1109/ISCIS.2009.5291847}
  {\path{doi:10.1109/ISCIS.2009.5291847}}.

\end{thebibliography}
\end{document}